\documentclass[
 amsmath,amssymb,
 aps, physrev,
]{revtex4-2}

\usepackage{graphicx}
\usepackage{dcolumn}
\usepackage{bm}
\usepackage{subcaption}
\usepackage{booktabs}
\usepackage{enumitem} 
\makeatletter
\newcommand{\setword}[2]{%
  \phantomsection
  #1\def\@currentlabel{\unexpanded{#1}}\label{#2}%
}
\makeatother

\begin{document}

\preprint{APS/Physical Review Fluids}

\title{\textbf{Physics-informed neural networks for phase-resolved data assimilation and prediction of nonlinear ocean waves} 
}%

\author{Svenja Ehlers}
\email[corresponding author: ]{svenja.ehlers@tuhh.de}
\affiliation{\footnotesize Dynamics Group, Hamburg University of Technology, Hamburg, Germany}

\author{Norbert Hoffmann}
\affiliation{\footnotesize Dynamics Group, Hamburg University of Technology, Hamburg, Germany}
\affiliation{\footnotesize Department of Mechanical Engineering, Imperial College London, London, United Kingdom}

\author{Tianning Tang}
\affiliation{\footnotesize Department of Engineering Science, University of Oxford, Oxford, United Kingdom}

\author{Adrian H. Callaghan}
\affiliation{\footnotesize Civil and Environmental Engineering, Imperial College London, London, United Kingdom}

\author{Rui Cao}
\affiliation{\footnotesize Civil and Environmental Engineering, Imperial College London, London, United Kingdom}

\author{Enrique M. Padilla}
\affiliation{\footnotesize BarcelonaTech, Laboratori d’Enginyeria Marítima, Universitat Politécnica de Catalunya, Barcelona, Spain}

\author{Yuxin Fang}
\affiliation{\footnotesize College of Oceanography and Space Informatics, China University of Petroleum, Qingdao, China}
\affiliation{\footnotesize Civil and Environmental Engineering, Imperial College London, London, United Kingdom}

\author{Merten Stender}
\affiliation{\footnotesize Cyber-Physical Systems in Mechanical Engineering, Technische Universität Berlin, Berlin, Germany}

\date{\today}

\begin{abstract}

The assimilation and prediction of phase-resolved surface gravity waves are critical challenges in ocean science and engineering. Potential flow theory (PFT) has been widely employed to develop wave models and numerical techniques for wave prediction. However, traditional wave prediction methods are often limited. For example, most simplified wave models have a limited ability to capture strong wave nonlinearity, while fully nonlinear PFT solvers often fail to meet the speed requirements of engineering applications. This computational inefficiency also hinders the development of effective data assimilation techniques, which are required to reconstruct spatial wave information from sparse measurements to initialize the wave prediction.
To address these challenges, we propose a novel solver method that leverages physics-informed neural networks (PINNs) that parameterize PFT solutions as neural networks. This provides a computationally inexpensive way to assimilate and predict wave data. The proposed PINN framework is validated through comparisons with analytical linear PFT solutions and experimental data collected in a laboratory wave flume. The results demonstrate that our approach accurately captures and predicts irregular, nonlinear, and dispersive wave surface dynamics. Moreover, the PINN can infer the fully nonlinear velocity potential throughout the entire fluid volume solely from surface elevation measurements, enabling the calculation of fluid velocities that are difficult to measure experimentally.

\end{abstract}


\maketitle


\section{Introduction}


Accurate knowledge of phase-resolved water wave dynamics is crucial in ocean engineering, for example, to assess the impact of waves on maritime structures. In most cases of nonlinear, phase-resolved ocean wave modelling, the problem is reduced to potential flow theory (PFT), governed by the Laplace differential equation and boundary conditions at the water surface and seabed. As the PFT cannot be solved analytically, various wave theories have emerged, using simplifying assumptions to derive analytical solutions, such as linear wave theory \citep{Airy1845}, Stokes weakly nonlinear wave theory \citep{Stokes1847} or the nonlinear Schrödinger equation \citep{Zakharov1968, Hasimoto1972}. 
In addition, numerical simulation methods are employed for modelling ocean waves based on the PFT: For example, boundary element methods (BEM) can be applied for linear problems in engineering applications with sufficiently fast computation. However, modelling nonlinear wave dynamics with BEM \citep{Grilli1989, Grilli2001} becomes computationally inefficient for large computational domains \citep{Papillion2020, Harris2014}. In contrast, the pseudo-spectral high-order spectral method (HOSM), independently developed by Dommermuth and Yue \cite{Dommermuth1987} and West et al. \cite{West1987}, is capable of handling waves in larger domains and provides the highest prediction accuracy across highly nonlinear wave conditions \citep[cf.][]{Wu2004, Blondel2010, Blondel2013, Klein2020}. Moreover, the HOSM is computationally efficient, provided that initial conditions are available \citep[cf.][]{Ducrozet2016, Koellisch2018, Klein2020}. 

However, spatio-temporal initial wave fields are rarely available in reality, as wave measurements are often sparse in space or time. This limitation has already been addressed by Padilla et al. \cite{Padilla2023} for certain one-dimensional laboratory conditions using spatial interpolation. A more general approach typically involves data assimilation methods to integrate observational data with the numerical model. In such methods, the wave prediction process typically consists of three steps:

\begin{enumerate}
\setlength{\itemsep}{0pt}
    \item[(\setword{S1}{step:1})] measurement of sparse wave-related quantities, e.g. by wave buoys, radar systems, optical measurements, etc.
    \item[(\setword{S2}{step:2})] reconstruction of the underlying, spatio-temporal wave field by assimilation methods to initialize a wave model 
    \item[(\setword{S3}{step:3})] forward propagation of the reconstructed wave field in time
\end{enumerate}

Assimilation methods often rely on variational optimization techniques to minimize a cost function that quantifies the agreement between wave surface measurements and the solution parameterized by a wave model \citep[cf.][]{Wu2004, Aragh2008, Blondel2010, Qi2018, Fujimoto2020, Desmars2020}.  Solving this inverse problem requires multiple forward evaluations to iteratively update model parameters, resulting in substantial computational costs for assimilation (\ref{step:2}) of nonlinear waves and thus hindering their real-time prediction (\ref{step:3}). Additionally, assimilation techniques often focus on aligning with elevation measurements only, neglecting the fundamental physics of surface gravity waves described by surface elevation and velocity potential. Although the nonlinear assimilation methods developed by Köllisch et al. \cite{Koellisch2018} and Desmars et al. \cite{Desmars2022, Desmars2023} can estimate a consistent surface velocity potential, they provide limited information in the vertical direction. Moreover, the accuracy of current assimilation and prediction approaches is constrained by the accuracy of the chosen wave model or numerical scheme, which is ultimately constrained by the computational speed during such iterative processes.

Recent studies have also used machine learning approaches to model waves following the \textit{Universal Approximation Theorem} \citep{Hornik1989, Cybenko1989}, which states that neural networks (NNs) with sufficient capacity can theoretically approximate any continuous function of arbitrary nonlinearity. As velocity potential $\Phi(x,t,z)$ and surface elevation $\eta(x,t)$ are nonlinear functions of space and time, this theorem provides strong motivation to approximate solutions to potential flow theory using NNs. 
Although recently applied in phase-resolved ocean wave prediction \citep[cf.][]{Law2020, Duan2020, Kagemoto2020, Mohaghegh2021, Klein2022, Zhang2022, Liu2022, Wedler2023, Chen2023, Feng2024} and reconstruction \citep[cf.][]{Ehlers2023, Zhao2023}, the success of these supervised machine learning approaches depends on the quality and quantity of labelled training data, making them likely to fail when wave conditions differ significantly from the training data. For instance, neural networks trained on synthetic data from numerical solvers may struggle when applied to real ocean data, but gathering real ocean training data is also challenging. While input data such as radar images $\xi(x,t)$ or buoy measurements $\eta(t)$ are accessible, obtaining wave elevation outputs $\eta(x,t)$ requires an impractically dense array of measurement buoys. Moreover, measurements can be influenced by noise and typically cover elevation-related quantities only, neglecting the velocity potential. This can lead to NN-based solutions that lack physical consistency for two reasons - their black-box nature, which relies on the properties of the training data, and their limited consideration of physical principles beyond surface elevation.

In light of these limitations, and given the prerequisite of addressing an inverse assimilation problem (\ref{step:2}) before approaching the forward prediction (\ref{step:3}), the emerging paradigm of physics-informed neural networks (PINNs) \citep{Raissi2019} offers a promising solution to enhance the reliability of phase-resolved wave prediction. PINNs are advantageous in handling ill-posed inverse problems by integrating sparse or noisy observational data with governing physical laws, typically formulated as partial differential equations (PDEs) \citep{Karniadakis2021}. The solution to this PDE is parameterized by a neural network, with the loss function accounting for the residual of the PDE, initial and boundary conditions and sparse system measurements, evaluated at collocation points within the computational domain. Leveraging automatic differentiation (AD) \citep{Baydin2018} to compute the differential operators in the residuals allows PINN solvers to operate without a numerical grid, offering an advantage over traditional numerical methods.

In recent years, PINNs have gained attention across various scientific disciplines, where solving forward, inverse and assimilation problems is in demand \citep{Cuomo2022}. These broad application possibilities include fluid dynamics \citep{Mao2020, Raissi2020, Cai2021a, vonSaldern2022}, acoustics \citep{RashtBehesht_2022}, heat transfer \citep{Zobeiry2021, Cai2021}, biomedical engineering \citep{Kissas2020, Sahli2020} climate modeling \citep{Kashinath2021} and material modelling \citep{Nguyen-Thanh2020,Goswami2020}.
PINNs have also been applied to surface wave-related research in shallow water regimes. For example, Wang et al.\cite{Wang2022} and Chen et al. \cite{Chen2023b} integrated residuals of the wave energy balance equation and linear dispersion to reconstruct near-shore wave heights and solve depth inversion problems. Other recent studies downscaled river models by integrating PINNs with the Saint-Venant equation \citep{Feng2023}, resolved assimilation problems by incorporating the Serre-Green-Naghdi equations \citep{Jagtap2022} or modelled flood-wave propagation over varying bathymetry using the shallow water equations \citep{Leiteritz2021}.

For the intermediate to deep water regime, we demonstrated that PINNs facilitate wave data assimilation by utilizing the nonlinear Schrödinger equation \citep{Ehlers2024}. However, the approach is inherently constrained by assumptions of narrow bandwidth and moderate wave steepness, which limit its applicability to realistic sea states characterized by broad-banded and irregular wave conditions. Another recent research captures irregular waves by incorporating the closed-form linear wave theory solution and demonstrates good agreement \citep{Liu2024}. This method, on the other hand, relies on a predefined solution rather than allowing the neural network to discover the solution to the PDEs, which is slightly on the edge of the original PINN paradigm of Raissi et al. \cite{Raissi2019}, and also limits the flexibility of the solutions. Therefore, further extensions are required for nonlinear waves, which may be addressed by directly solving the fully nonlinear potential flow equations. Concurrent with our research on the PFT-PINN, Lu et al. \cite{Lu2024} independently developed a similar, initial approach for this purpose. Their method, evaluated on synthetic data in shallow and intermediate depths, introduces a normalization strategy based on wavenumber and frequency. While effective in certain conditions, this approach seems primarily suited to narrow-banded wave scenarios in limited depths and may require further development to ensure applicability to realistic sea states. Therefore, we aim to develop a PFT-PINN solver with broad applicability for assimilation and prediction of various wave conditions, even though this is challenging compared to most PINN applications due to the
\begin{itemize}
\setlength{\itemsep}{0pt}
    \item [(\setword{C1}{chal:1})]search for two solutions fields simultaneously - the potential field $\Phi(x,t,z)$ and the surface elevation $\eta(x,t)$ 
    \item [(\setword{C2}{chal:2})] dependence of the two solutions on different domains and a different number of input variables
    \item [(\setword{C3}{chal:3})] requirement of fulfilling boundary conditions at the changing, instantaneous free surface $z=\eta(x,t)$
    \item [(\setword{C4}{chal:4})] availability of sparse measurements for elevation only ($\eta_\mathrm{m}$) but not for the potential (ill-posed inverse problem)
\end{itemize}
However, after successfully designing and training a framework that addresses all these challenges, we hypothesize that the resulting PFT-PINN solver is capable of 
\begin{itemize}
\setlength{\itemsep}{0pt}
    \item [(\setword{H1}{hyp:1})] reconstructing broad-banded waves with irregular shapes 
    \item [(\setword{H2}{hyp:2})] inferring the physically consistent potential $\Phi(x,t,z)$ while reconstructing the elevation $\eta(x,t)$ from sparse surface measurements $\eta_\mathrm{m}(t)$ only
    \item [(\setword{H3}{hyp:3})] reconstructing waves of not predefined, theoretically arbitrary nonlinear order 
    \item [(\setword{H4}{hyp:4})] jointly solving assimilation (\ref{step:2}) and prediction (\ref{step:3}) tasks within the same optimization problem
\end{itemize}

We proceed as follows: Section~\ref{sec:PFT} introduces the potential flow equations of ocean gravity waves, while Section~\ref{sec:wave_tank_experiments} describes the set-up and data structure of the wave flume measurements. Section~\ref{sec:Method} presents the developed PINN framework integrating the potential flow equations, which is verified in Section~\ref{sec:Results} by solving wave assimilation and prediction problems using sparse synthetic waves and wave flume measurements. We summarize our findings and discuss the opportunities for using PINNs in real-world scenarios in Section~\ref{sec:conclusion}.

\section{Potential flow theory}
\label{sec:PFT}

Assuming water to be a Newtonian fluid, which is homogeneous, incompressible, inviscid and irrotational, the flow field is expressed as a velocity potential $\Phi(x,z,t)$ satisfying the Laplace equation
\begin{align}
    \nabla^2 \Phi = \Phi_{xx} + \Phi_{zz}&=0  \label{eq:Laplace}
\end{align}
within the fluid domain $\Omega$. This domain is limited by the boundary conditions (BCs)
\begin{align}
    \eta_t +  \eta_x \Phi_x - \Phi_z&=0 \hspace{0.5cm}      \text{on } z=\eta(x, t)  \label{eq:kinBC}\\ 
    \Phi_t +g \eta   + \frac{1}{2} \left(\Phi_{x}^2 +\Phi_{z}^2  \right)&=0 \hspace{0.5cm} \text{on } z=\eta(x, t) \label{eq:dynBC}\\
    \Phi_z &=0  \hspace{0.5cm}  \text{on } z=-d. \label{eq:bot} 
\end{align}
The kinematic surface BC (Eq.~\eqref{eq:kinBC}) prevents water particles from leaving the free wave surface $\eta(x,t)$, while the dynamic surface BC (Eq.~\eqref{eq:dynBC}) ensures constant dynamic pressure at the free surface. Here, $g$ represents the gravitational acceleration. The third BC (Eq.~\eqref{eq:bot}) establishes a rigid and impermeable sea bottom at depth $d$.  

Although the Laplace equation is a linear PDE, solving the complete problem (Eqs. \eqref{eq:Laplace}-\eqref{eq:bot}) is challenging, e.g. due to the nonlinear surface BCs that need to be satisfied at the instantaneous free surface. Solutions can be approximated by numerical procedures or closed-form solutions can be derived, such as linear wave theory (LWT) \citep{Airy1845}. The LWT will be applied in this work to provide an analytical solution for the validation of the developed PINN approach.

The linearization of the surface BCs yields analytic equations for $\Phi(x,z,t)$ and $\eta(x,t)$ \citep{Chakrabarti1987} and superposition principle allows to model irregular water waves as a Fourier series of $N$ components with different amplitudes $a_n$, frequencies $\omega_n$, wavenumbers $k_n$ and phase shifts $\varphi_n$ as
\begin{align}
    \Phi(x,z,t)&= \sum_{n=1}^N \frac{g a_n}{\omega_n} \cdot \frac{\cosh\left(k_n(z+d)\right)}{\cosh(k_nd)} \cdot \sin(k_n x -\omega_n t + \varphi_n)\label{eq:LWT_phi}\\
    \eta(x,t) &= \sum_{n=1}^N a_n \cdot \cos(k_n x -\omega_n t + \varphi_n), \label{eq:LWT_eta}
\end{align}
whereas $k_n$ and $\omega_n$ are constrained by the linear dispersion relation
\begin{equation}
    \omega_n = \sqrt{g k_n\cdot \tanh (k_n d)}. \label{eq:linear_dispersion}
\end{equation}

\section{Wave flume experiments}
\label{sec:wave_tank_experiments}

\begin{figure}[ht!]
     \centering
    \begin{subfigure}[b]{0.8\textwidth}
         \centering
         \begin{scriptsize}
         \includegraphics[width=0.6\textwidth]{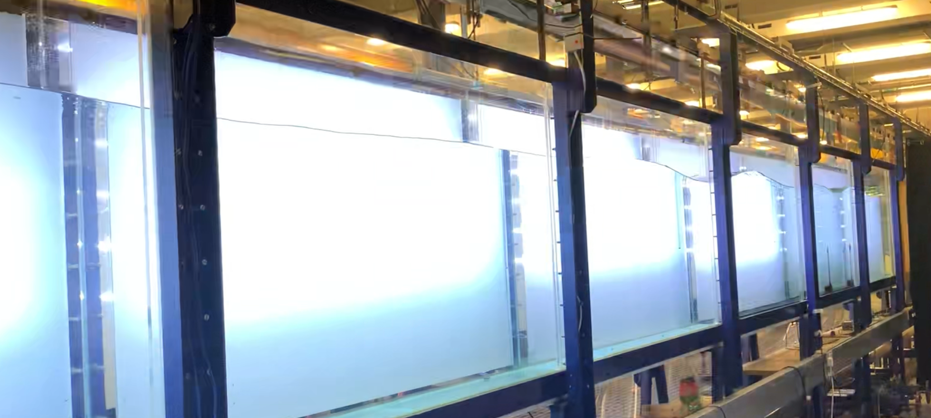}
         \end{scriptsize}
         \caption{}
         \vspace{0.25cm}
         \label{subfig:photo}
     \end{subfigure}
     \begin{subfigure}[b]{0.48\textwidth}
         \centering
         \begin{scriptsize}
         \includegraphics[width=\textwidth]{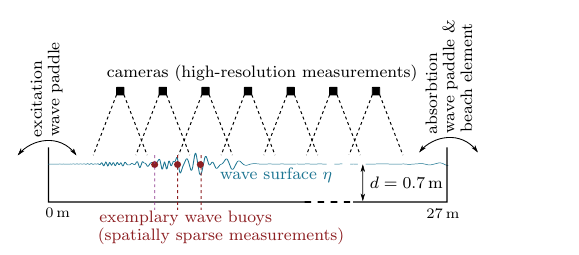}
         \end{scriptsize}
         \vspace{0.75cm}
         \caption{}
         \label{subfig:wavetank}
     \end{subfigure}
     \hspace{0.4cm}
     \begin{subfigure}[b]{0.45\textwidth}
         \centering
         \includegraphics[width=0.9\textwidth]{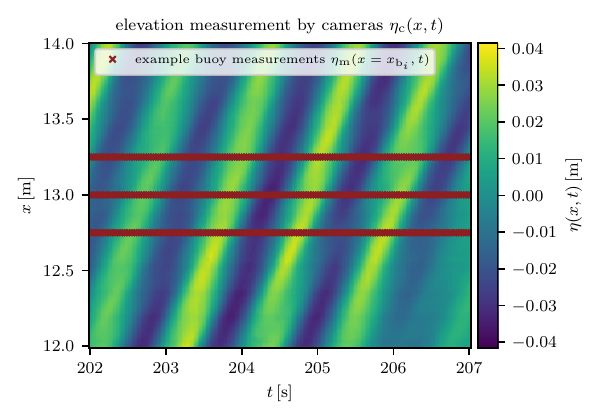}
         \caption{}
         \label{subfig:measurement_visualization}
     \end{subfigure}
     \caption{In the laboratory wave flume (a), lateral cameras capture the wave elevation with high resolution (b), resulting in a spatio-temporal data structure after preprocessing (c) that provides a ground truth ($\eta_\mathrm{true}$) for comparing the PINN reconstruction ($\tilde{\eta}$) between sparse buoy time series data ($\eta_\mathrm{m}$).}
     \label{fig:SetupMeasurement}
\end{figure}

We further evaluate our PFT-PINN approach using real measurements collected from laboratory experiments to validate its performance beyond the analytical LWT solution. This experimental data is a subset of a larger experimental campaign, MARATHON, which concerns unidirectional random sea states. The experiments were conducted in 2021 in a glass-walled wind-wave flume (Fig. \ref{subfig:photo}) located at the Hydrodynamics laboratory at the Civil and Environmental Engineering Department of Imperial College London. The wave flume has a length of $27\, \mathrm{m}$, a width of $0.3\, \mathrm{m}$, and  a water depth of $d=0.7\, \mathrm{m}$. The paddle at one end of the flume was responsible for wave generation, while the opposite paddle, along with a beach element, was used to dissipate incident wave energy and minimise reflections. This configuration has been demonstrated to provide satisfying wave energy dissipation efficiency with maximally only 3\% of energy being reflected \citep{Cao2023}.

The MARATHON random waves were generated based on a JONSWAP-type spectrum and designed following the NewWave theory \citep{Hasselmann1973, Tromans1991}.  The wave spectrum was characterized by a peak enhancement factor of $\gamma=3$ and a peak wave period of $T_\mathrm{p}=1.2\, \mathrm{s}$, corresponding to a peak frequency $\omega_\mathrm{p}=5.236 \, \tfrac{\mathrm{rad}}{\mathrm{s}}$, peak wavelength $L_\mathrm{p}=2.171\, \mathrm{m}$ and peak wavenumber $k_\mathrm{p}=2.894\, \tfrac{\mathrm{rad}}{\mathrm{ms}}$ for the given water depth, assuming linear dispersion relation applies. We selected a sequence of non-breaking random waves from the MARATHON dataset that are specified by their target wave steepness, $\epsilon=0.5H_\mathrm{s} k_\mathrm{p}$, where $H_\mathrm{s}$ represents the significant wave height. The wave steepness values increased from $\epsilon=0.05$ to $\epsilon=0.11$ in increments of 0.02, representing increasing wave field nonlinearity. 

These random waves were imaged at a frame rate of 32 Hz by seven CCD (charge-coupled device) cameras mounted laterally that captured the wave profile through the flume glass sidewalls. The surface elevations were extracted from the imagery data using the CMG method developed in Cao et al. \cite{Cao2024}. To fill the missing regions between the fields of view of adjacent cameras, interpolation was performed using the S-interp function developed in Padilla et al. \cite{Padilla2023}. As a result, a continuous 9-meter span of surface elevation data was obtained with a spatial resolution of $\Delta x= 0.001 \, \mathrm{m}$ and a temporal resolution of $\Delta t = 0.03125 \, \mathrm{s}$. These high-resolution spatio-temporal measurements far surpass the typical resolution of most experimental set-ups and oceanographic wave observations, which are usually limited to time-series measurements at sparse points, as exemplified by wave rider buoys in Fig. \ref{subfig:wavetank}. 

To explore the applicability of our PINN approach in real-world scenarios, we select only a few discrete spatial points ($x_{\mathrm{wb},j}$) as hypothetical buoy measurements. The spatial extent of the wave field between these buoy measurements ($\eta_\mathrm{m}$) is to be reconstructed using the PFT-PINN. The surface elevations obtained from image processing serve as the ground truth ($\eta_\mathrm{true}$) for comparison with the reconstructed wave field, which is denoted as $\Tilde{\eta}$ in the following. The relationship between camera-derived measurements and hypothetical buoy measurements is illustrated in Fig.~\ref{subfig:measurement_visualization}. \\

\section{PINN Method}
\label{sec:Method}

In the following, we introduce the PINN framework designed to address the challenges \ref{chal:1}-\ref{chal:4} associated with solving the potential flow equations (Eqs. \eqref{eq:Laplace}-\eqref{eq:bot}). Besides the network architecture, we define the computational domain using various sets of collocation points and describe the method for hard-encoding sparse surface measurements into the network design. We present the multi-objective optimization method for training, followed by a metric to compare PINN reconstructions and predictions with a ground truth.

\subsection{PINN architecture}

\begin{figure}[ht]
\vspace{0.2cm}
\centering
\includegraphics[width=0.97\textwidth]{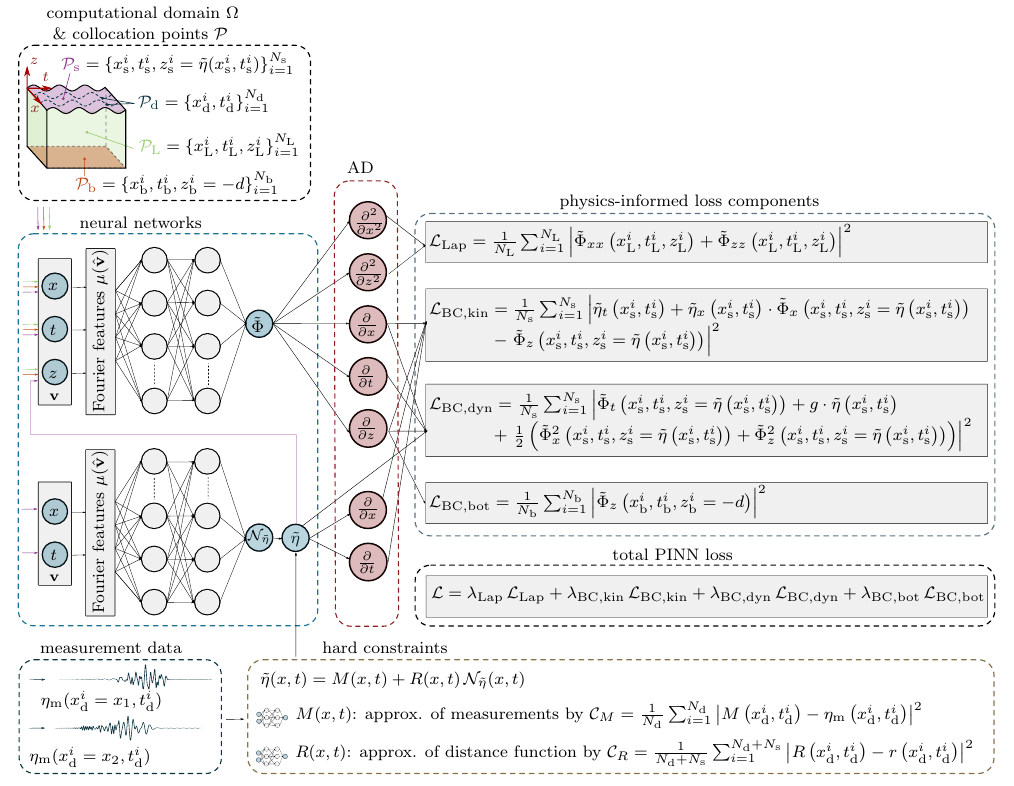}
\caption{ Physics-informed neural network (PINN) framework to solve the potential flow theory (PFT) of ocean gravity waves. Two neural networks run in parallel to approximate the solutions of the velocity potential $\tilde{\Phi}(x,t,z)$ and surface elevation $\tilde{\eta}(x,t)$. Collocation points at the instantaneous free surface $(\mathcal{P}_\mathrm{s})$, seabed $(\mathcal{P}_\mathrm{b})$ and in the fluid domain $(\mathcal{P}_\mathrm{L})$ serve as inputs $\mathbf{v}$ for the NNs, are normalized to $\hat{\mathbf{v}}$ and lifted to a higher-dimensional feature space by Fourier embedding $\mu(\hat{\mathbf{v}})$. Surface measurements $\eta_\mathrm{m}$ at sparse points $(\mathcal{P}_\mathrm{d})$ hard-constrain the direct output of the elevation network $\mathcal{N}_{\tilde{\eta}}$ with a smooth distance function $R(x,t)$ and extension $M(x,t)$ of $\eta_\mathrm{m}$, both approximated from pre-trained low-capacity NNs. Parametrizing the PFT-solutions as neural networks $\tilde{\eta}(x,t)$ and $\tilde{\Phi}(x,t,z)$ enables continuous differentiation, allowing for computation of loss components using automatic differentiation (AD). These loss components include residuals of the Laplace equation $(\mathcal{L}_\mathrm{Lap})$, and boundary conditions $(\mathcal{L}_\mathrm{BC,kin}, \, \mathcal{L}_\mathrm{BC,dyn}, \, \mathcal{L}_\mathrm{BC,bot})$. The total loss $\mathcal{L}$ is minimized during training to approximate $\tilde{\eta}(x,t)$ and $\tilde{\Phi}(x,t,z)$.}
\label{fig:PINN}
\end{figure}

The PFT-PINN first requires a computational domain for one-dimensional wave propagation, which is illustrated on top of Fig. \ref{fig:PINN} and defined in Cartesian coordinates as 
\begin{equation}
\Omega= \big\{(x,t,z): x \in \big[x_0, x_\mathrm{max}\big]\mathrm{m}, \, t \in \big[t_0, t_\mathrm{max}\big]\mathrm{s}, \, z \in \big[-d, z_\mathrm{max}\big] \mathrm{m} \big\}.
\end{equation}
The origin of the coordinate system is located on the still water level $z=0$, with the $z$-axis pointing upwards, while the upper limit of $\Omega$ is a dynamic boundary defined by the instantaneous free surface $z_\mathrm{max}=\eta(x,t)$.

In contrast to typical PINN applications, our scenario requires the simultaneous computation of two solution fields: the surface elevation $\tilde{\eta}(x,t)$ pertaining solely to the spatio-temporal fluid surface and the velocity potential $\tilde{\Phi}(x,t,z)$ integrating additional depth information (\ref{chal:1}-\ref{chal:2}). Two neural networks are employed in parallel to approximate these fields, as depicted in Figure \ref{fig:PINN}. Empirical analysis revealed that four hidden layers, each containing 200 neurons, are sufficient to model both solution fields accurately. The $\tanh$-activation function is used and the network weights are initialized by Xavier initialization \citep{Glorot2010}, with biases set to zero. Moreover, all input points $\mathbf{v}=(x^i, t^i)$ for $\tilde{\eta}$ or $\mathbf{v}=(x^i, t^i, z^i)$ for $\tilde{\Phi}$ are normalized to $\mathbf{\hat{v}} \in [-1,1]$ for each dimension and mapped to a higher-dimensional Fourier feature space. This transformation is proposed by \cite{Tancik2020} and defined as 
\begin{equation}
    \mu(\hat{\mathbf{v}}) = \left[\sin(2 \pi \mathbf{F} \hat{\mathbf{v}}); \, \cos(2 \pi \mathbf{F} \hat{\mathbf{v}}); \hat{\mathbf{v}} \right]^T.
    \label{eq:FFembedding}
\end{equation}
It allows to represent complex functions more effectively by capturing a wider range of frequencies and thus helps to mitigate the spectral bias problem, which is the tendency of NNs to prefer converging to the lower-frequency solution components \citep{Rahaman2019}. The entries of $\mathbf{F} \in \mathbb{R}^{n_\mathrm{f} \times n_\mathrm{v} }$ are trainable parameters and initialized with $n_\mathrm{f}=10$ evenly spaced values in $[-0.4, 0.4]$, while $n_\mathrm{v}$ is the number of input coordinates of the respective network. 
Thus, the chosen normalization and embedding approach allows the network to capture realistic, broad-banded sea states and complex multi-scale behaviour for irregular wave dynamics. 

\subsection{Definition of collocation points}

To approximate the solution to the potential flow theory, the PINN requires various sets of collocation points $\mathcal{P}$, which are specific sample points in the computational domain $\Omega$ used to evaluate the residuals of governing equations. These collocation points serve as inputs $ \mathbf{v}$ to the NNs for computing the outputs $\Tilde{\eta}(x,t)$ and $\Tilde{\Phi}(x,t,z)$, that are employed for calculating spatial and temporal derivatives with respect to input variables using automatic differentiation (AD) \citep{Baydin2018}. This is essential for evaluating the loss components as the total loss function is defined as the sum of the mean-squared errors (MSEs) from the residuals of the Laplace equation ($\mathcal{L}_\mathrm{Lap}$), kinematic surface, dynamic surface and bed boundary conditions ($\mathcal{L}_\mathrm{BC,kin}$, $\mathcal{L}_\mathrm{BC,dyn}$, $\mathcal{L}_\mathrm{BC,bot}$):

\begin{footnotesize}
\begin{align}
    \mathcal{L}_\mathrm{BC,bot} &= \frac{1}{N_\mathrm{b}} \sum_{i=1}^{N_\mathrm{b}} \left| \tilde{\Phi}_z  \left( x_\mathrm{b}^{i},t_\mathrm{b}^{i}, z_\mathrm{b}^{i}=-d \right) \right|^2 \label{eq:loss bot} \\
    \mathcal{L}_\mathrm{BC,kin} &=\frac{1}{N_\mathrm{s}} \sum_{i=1}^{N_\mathrm{s}}  \left| \tilde{\eta}_t \left(x_\mathrm{s}^{i},t_\mathrm{s}^{i}\right) + \tilde{\eta}_x \left(x_\mathrm{s}^{i},t_\mathrm{s}^{i}\right) \cdot \tilde{\Phi}_x\left(x_\mathrm{s}^{i},t_\mathrm{s}^{i}, z_\mathrm{s}^{i}=\tilde{\eta}\left(x_\mathrm{s}^{i},t_\mathrm{s}^{i}\right) \right) - \tilde{\Phi}_z \left(x_\mathrm{s}^{i},t_\mathrm{s}^{i}, z_\mathrm{s}^{i}=\tilde{\eta}\left(x_\mathrm{s}^{i},t_\mathrm{s}^{i}\right) \right) \right| ^2 \label{eq:loss kin}\\ 
    \mathcal{L}_\mathrm{BC,dyn} &= \frac{1}{N_\mathrm{s}} \sum_{i=1}^{N_\mathrm{s}} \left|\tilde{\Phi}_t\left(x_\mathrm{s}^{i},t_\mathrm{s}^{i}, z_\mathrm{s}^{i}= \tilde{\eta} \left(x_\mathrm{s}^{i},t_\mathrm{s}^{i}\right)  \right) +g \tilde{\eta}\left(x_\mathrm{s}^{i},t_\mathrm{s}^{i}\right) +  \frac{1}{2} \left(\tilde{\Phi}_x^2\left(x_\mathrm{s}^{i},t_\mathrm{s}^{i}, z_\mathrm{s}^{i}=\tilde{ \eta}\left(x_\mathrm{s}^{i},t_\mathrm{s}^{i}\right) \right) +\tilde{\Phi}_z^2 \left(x_\mathrm{s}^{i},t_\mathrm{s}^{i},  z_\mathrm{s}^{i}=\tilde{\eta}\left(x_\mathrm{s}^{i},t_\mathrm{s}^{i}\right) \right) \right)\right|^2\label{eq:loss dyn}\\
    \mathcal{L}_\mathrm{Lap} &= \frac{1}{N_\mathrm{L}} \sum_{i=1}^{N_\mathrm{L}} \left|\tilde{\Phi}_{xx} \left(x_\mathrm{L}^i,t_\mathrm{L}^i, z_\mathrm{L}^i\right)+ \tilde{\Phi}_{zz}\left(x_\mathrm{L}^i,t_\mathrm{L}^i, z_\mathrm{L}^i\right)\right|^2. \label{eq:loss Laplace} 
\end{align}
\end{footnotesize}

Empirically, we found that defining $N_\mathrm{b}=1,000$ collocation points at the seabed, denoted as $\mathcal{P}_\mathrm{b}=\{x_\mathrm{b}^{i}, t_\mathrm{b}^{i}, z_\mathrm{b}^{i}=-d\}_{i=1}^{N_\mathrm{b}}$, is sufficient for evaluating $\mathcal{L}_\mathrm{BC,bot}$. For $\mathcal{L}_\mathrm{BC,kin}$ and $\mathcal{L}_\mathrm{BC,dyn}$, another set of $N_\mathrm{s} = 5,000$ collocation points is defined as $\mathcal{P}_\mathrm{s}=\{x_\mathrm{s}^{i}, t_\mathrm{s}^{i}, z_\mathrm{s}^{i}=\Tilde{\eta}(x_\mathrm{s}^{i}, t_\mathrm{s}^{i})\}_{i=1}^{N_\mathrm{s}}$, where the $ z_\mathrm{s}^{i}$-values are not fixed but adhere to the instantaneous free surface (\ref{chal:3}) in the original Cartesian coordinates by linking the $\Tilde{\eta}$-network output to the $z$-input of the $\Tilde{\Phi}$-network, as shown in Fig. \ref{fig:PINN}. For $\mathcal{L}_\mathrm{Lap}$, we sample $N_\mathrm{L}=30,000$ collocation points $\mathcal{P}_\mathrm{L}=\{x_\mathrm{L}^{i}, t_\mathrm{L}^{i}, z_\mathrm{L}^i\}_{i=1}^{N_\mathrm{L}}$ across the entire $\Omega$ and extend the initial range of $z_\mathrm{L}^{i}$ to $ [-d, \, 1.1\cdot \max(\eta_\mathrm{m})]$ to account for the fluid volume directly underneath the free instantaneous surface elevation $\eta(x,t)$ but above the still water level. The condition
\begin{equation}
    z_\mathrm{L}^i = \begin{cases}
z_\mathrm{L}^i & \text{if} \; z_\mathrm{L}^i < \tilde{\eta}(x_\mathrm{L}^{i},t_\mathrm{L}^{i})\\
\tilde{\eta}(x_\mathrm{L}^{i},t_\mathrm{L}^{i}) & \text{if} \; z_\mathrm{L}^i \ge \tilde{\eta}(x_\mathrm{L}^{i},t_\mathrm{L}^{i})
\end{cases}
\end{equation}
then ensures that the $z_\mathrm{L}^{i}$-value remains in the valid fluid domain, as it changes during each training epoch.

\subsection{Hard constraints for elevation measurements}
\label{sec:hard_constraints}
To guide the PINN solution away from trivial outcomes, sparse measurements of the specific wave instance are required alongside Eqs. \eqref{eq:loss bot}-\eqref{eq:loss Laplace}. Wave measurement data $\eta_\mathrm{m}$, typically provided as snapshots or time-series at discrete data points $\mathcal{P}_\mathrm{d}=\{x_\mathrm{d}^{i}, t_\mathrm{d}^{i}\}_{i=1}^{N_\mathrm{d}}$, may be incorporated as \textit{soft constraint} by adding an additional penalty term ($\mathcal{L}_\mathrm{data}$) to the total loss function. However, manual tuning of weighting factor $\lambda_\mathrm{d}$ for $\mathcal{L}_\mathrm{data}$ often leads to poor learning efficiency \citep{Cuomo2022}. To overcome this, we implement a \textit{hard constraint} approach for the PFT-PINN, directly embedding data constraints into the network architecture rather than treating them as an additional loss component \citep{BERG2018, Lu2021}. This hard encoding is accomplished by modifying the direct neural network output $\mathcal{N}_{\tilde{\eta}}(x,t)$ according to
\begin{equation}
    \tilde{\eta}(x,t) = M(x,t) + R(x,t) \, \mathcal{N}_{\tilde{\eta}}(x,t).
\end{equation}
Here, $M(x,t)$ is a smooth extension of the measurement data $\eta_\mathrm{m}$, while $R(x,t)$ is a smooth distance function derived from the minimum spatio-temporal distance from the surface points $\mathcal{P}_\mathrm{s}$ to the measurement points $\mathcal{P}_\mathrm{d}$, denoted as $r(x,t)$. Both, $M(x,t)$ and $R(x,t)$, are normalized within $[0,1]$ and approximated using a low-capacity NN with two hidden layers containing 50 neurons, by minimizing the cost functions $\mathcal{C}_\mathrm{M}$ and $\mathcal{C}_\mathrm{R}$ defined in the lower-right of Fig. \ref{fig:PINN}. After this pre-training, the parameters of $M(x,t)$ and $R(x,t)$ are frozen to hard-constrain the subsequent PFT-PINN training process.
The absence of measurements for the potential field restricts hard constraints to the elevation network, making the problem highly ill-posed \citep{Jagtap2022}: While elevation measurements guide the solution towards the targeted instance, the potential must be inferred solely through the coupling of $\Tilde{\eta}$ and $\Tilde{\Phi}$ in $\mathcal{L}_\mathrm{BC,kin}$ and $\mathcal{L}_\mathrm{BC,dyn}$ (\ref{chal:4}).

\subsection{Multi-objective optimization procedure}
\label{sec:multi-objective optimization}
During each training epoch, all collocation points ($ \mathcal{P}_\mathrm{s}, \mathcal{P}_\mathrm{b}$ and $\mathcal{P}_\mathrm{L}$) are passed through the $\tilde{\eta}$- and $\tilde{\Phi}$ network to evaluate the loss components (Eqs. \eqref{eq:loss bot}-\eqref{eq:loss Laplace}). This creates a multi-objective optimization problem with four potentially conflicting components $\mathcal{L}_i$, which may also conflict with the hard constraints of the measurement data due to model or data inaccuracies. Furthermore, these components can vary significantly in magnitude due to differences in physical units and network initialization. To avoid gradient pathologies arising from imbalanced loss terms \citep{Wang2020}, selecting appropriate weighting factors $\lambda_i$ is crucial to compute the total loss as a weighted sum of the $m$ components
\begin{equation}
    \mathcal{L} = \lambda_\mathrm{Lap} \cdot \mathcal{L}_\mathrm{Lap} + \lambda_\mathrm{BC,kin} \cdot \mathcal{L}_\mathrm{BC,kin} + \lambda_\mathrm{BC,dyn} \cdot \mathcal{L}_\mathrm{BC,dyn} + \lambda_\mathrm{BC,bot} \cdot \mathcal{L}_\mathrm{BC,bot}. \label{eq:total_loss}
\end{equation}
As manual tuning of $\lambda_i$ requires extensive grid-search procedures, we use the self-adaptive loss balancing algorithm \textit{ReLoBRaLo} (Relative Loss Balancing with Random Lookbacks, \cite{Bischof2021}), updating each weighting factor $\lambda_i$ for the corresponding loss component $\mathcal{L}_i$ at each epoch ($e$) according to 
\begin{equation}
\begin{split}
    \lambda_i^{(e)}&=\alpha \left(\rho \lambda_i^{(e-1)} + (1- \rho) \nu_i^{(e;\,0)}\right)+ (1-\alpha) \nu_i^{(e; \,e-1)}\\
     \nu_i^{(e,\,e')} & = m \cdot \frac{\exp{\left(\frac{\mathcal{L}_i^{(e)}}{\tau \mathcal{L}_i^{(e')}}\right)}}{\sum\limits_{j=1}^m \exp{\left(\frac{\mathcal{L}_j^{(e)}}{\tau \mathcal{L}_j^{(e')}}\right)}}, \; \; i \in \{1, \hdots, m \}.
\end{split}
\end{equation}

\textit{ReLoBRaLo} dynamically adjusts loss weights $\lambda_i$ based on the relative magnitudes of the $m$ loss components $\mathcal{L}_i$ by combining benefits of several loss balancing strategies \citep[cf.][]{Chen2018, Heydari2019, Wang2020}. It normalizes the rate of change in each loss component between consecutive epochs using a softmax function. The parameter $\alpha$ balances historical weight information with new updates, where higher $\alpha$ emphasize historical weights and increase training stability. The temperature $\tau$ controls the sensitivity of weight assignments, where smaller values turn the balancing scheme more aggressive. The Bernoulli random variable $\rho$ with $\mathbb{E}[\rho]$ near 1 introduces random lookbacks $\nu_i^{(e;\,0)}$  to the initial loss value $\mathcal{L}_i^{(0)}$ to escape local minima. For the PFT-PINN, we empirically determined these values as $\alpha=0.95$, $\tau=20$ and $\mathbb{E}[\rho]=0.98$.

The loss function in PINNs is typically minimized using a two-step strategy incorporating an Adam and the L-BFGS optimizer \citep{Markidis2021}. For the total loss Eq. \eqref{eq:total_loss}, the AmsGrad modification of the Adam optimizer \citep{Kingma2014,Reddi2019} with a learning rate $lr = 0.0005$ is used for a predefined number of 5,000 epochs. Subsequently, the L-BFGS optimizer \citep{Liu1989} with a history size of 30 refines the small-scale solution components and is executed until either a termination criterion or a maximum of 40,000 epochs is reached.  Both optimizers are implemented in the PyTorch library \citep{Pytorch2019} and the training is performed on a NVIDIA GeForce RTX 3090 GPU, taking around 30 minutes.

\subsection{PINN evaluation metric}

The evaluation of the PFT-PINN's wave elevation $\tilde{\eta}$ or potential $\tilde{\Phi}$ after training requires a metric comparing them with a ground truth, $\eta_\mathrm{true}$ or $\Phi_\mathrm{true}$. While Euclidean distance-based metrics like MSE are scale-dependent and treat deviations in frequency and phase as amplitude errors \citep{Wedler2022, Iida2024}, the surface similarity parameter (SSP)  \citep{Perlin2014} integrates phase, amplitude, and frequency errors into a scalar value. For spatio-temporal 2D surfaces, the SSP is defined as
\begin{equation}
    \mathrm{SSP}(\mathbf{y}_\mathrm{true}, \Tilde{\mathbf{y}}) = \frac{\sqrt{\int | F_{\mathbf{y}_\mathrm{true}}(\omega, k)-F_{\Tilde{\mathbf{y}}}(\omega, k)|^2 d \omega d k}}{\sqrt{\int | F_{\mathbf{y}_\mathrm{true}}(\omega, k)|^2 d \omega d k}+\sqrt{\int |F_{\Tilde{\mathbf{y}}}(\omega, k)|^2 d \omega d k}} \in [0, 1],
    \label{eq:SSP}
\end{equation}
which can also be simplified for 1D slices in space or time. Here, $\omega$ denotes the wave frequency and $k$ the wavenumber vector of the discrete Fourier transforms $F_{\mathbf{y}_\mathrm{true}}$ and $F_{\Tilde{\mathbf{y}}}$. Thereby $\mathbf{y}_\mathrm{true}$ represents either the ground truth surface elevation $\eta_\mathrm{true}(x,t)$ or the sliced potential $\Phi_\mathrm{true}(x,t,z=z_j)$ at depth $z_j$, while $\Tilde{\mathbf{y}}$ refers to the PFT-PINN's outputs $\Tilde{\eta}(x,t)$ or $\Tilde{\Phi}(x,t,z=z_j)$. Moreover, the normalized error metric provides a straightforward error assessment, with $\mathrm{SSP}=0$ indicating perfect alignment and $\mathrm{SSP}=1$ implying a comparison against a zero or phase-inverted surface. Therefore, the SSP has been recently applied in ocean wave prediction and reconstruction \citep[cf.][]{Klein2020, Klein2022, Wedler2022, Wedler2023, Desmars2021, Desmars2022, Desmars2023, Luenser2022, Kim2023, Ehlers2023, Ehlers2024}.  Appendix Figure~\ref{fig:SSPs} illustrates the appearance of different SSP values by comparing two wave surfaces with increasing alignment.

\section{Results}
\label{sec:Results}

In the following, Subsection \ref{sec:assimilation} first demonstrates a data assimilation task using the PFT-PINN, by estimating surface elevation and velocity potential at locations between spatially sparse elevation measurements. Particularly, we validate our PINN's ability to handle irregular sea states obtained synthetically from an analytical solution, before we utilize data stemming from wave flume experiments. Afterwards, we asses the PINN's capabilities in forecasting nonlinear wave evolution in Subsection~\ref{sec:prediction}. Consequently, the PINN is validated under full-scale wave conditions for the analytical wave data and is further demonstrated to be suitable for downscaled data from wave experiments, thus operating independently of the data scaling.

\subsection{Assimilation}
\label{sec:assimilation}

The initial phase of our analysis focuses on data assimilation, a technique to integrate partial observations with numerical models to solve an inverse problem, where the unknown parameter is the state of the dynamical system \citep{Sanz-Alonso2023}. In ocean engineering, assimilation typically involves reconstructing a wave field $\tilde{\eta}(x,t)$ (the state) from sparse measurements (the observations), such as snapshots $\eta_\mathrm{m}(x, t=t_{\mathrm{sn},k})$ or wave buoy time series $\eta_\mathrm{m}(x=x_{\mathrm{wb},j}, t)$. In our study, the wave field is parameterized by the PFT-PINN (the numerical model) and the training procedure integrates it with the measurements $\eta_\mathrm{m}$ and the underlying physical constraints. However, the availability of sparse measurements for only one quantity of interest requires deducing the potential $\tilde{\Phi}$ entirely from the surface elevation $\tilde{\eta}$. Section~\ref{sec:LWT_assimilation} demonstrates the assimilation property using a wave surface and potential from linear wave theory. Section~\ref{sec:tank_assimilation} extends this to a nonlinear irregular wave field generated in a laboratory wave flume.

\subsubsection{Linear irregular waves in deep water}
\label{sec:LWT_assimilation}
First, the PFT-PINN's assimilation capability is assessed using an analytical solution for wave surface elevation $\eta_\mathrm{true}(x,t)$ obtained via superposition of three linear components (Eq.~\eqref{eq:LWT_eta}) with parameters listed in Table~\ref{tab:params_LWT}. The use of analytical solutions has the advantage that they additionally provide a ground truth potential $\Phi_\mathrm{true}(x,t,z)$ (Eq.~\eqref{eq:LWT_phi}) for evaluating the PINN's solution. However, to emulate realistic scenarios, we assume that time-series measurements $\eta_\mathrm{m}(x=x_{\mathrm{wb},j},t)$ are available only at specific wave buoy positions, requiring the PINN to reconstruct the surface elevation in between $\tilde{\eta}(x,t)$ and the entire potential $\tilde{\Phi}(x,t,z)$. 
\begin{table}[ht!]
\centering
\caption{Parameters of the three wave components, superimposed according to linear wave theory in Eqs. \eqref{eq:LWT_phi}-\eqref{eq:LWT_eta}, to generate the analytic ground truth for elevation $\eta_\mathrm{true}(x,t)$ and potential $\Phi_\mathrm{true}(x,t,z)$. This data is used to extract synthetic surface measurements $\eta_\mathrm{m}(x=x_{\mathrm{wb},j},t)$ for PINN training and to validate the PFT-PINN solutions $\tilde{\eta}(x,t)$ and $\tilde{\Phi}(x,t,z)$ after training.}
\begin{tabular}{llcccccccc}
\toprule
\toprule
& \multicolumn{4}{c}{base parameters} & \multicolumn{4}{c}{resulting parameters}  \\
\cmidrule(rl){2-5} \cmidrule(rl){6-9}  
 component $i$ & $\omega_i\, [\frac{\mathrm{rad}}{\mathrm{s}}]$ & $a_i \, [\mathrm{m}]$ & $d_i\, [\mathrm{m}]$ & $\phi_i \, [\mathrm{rad}]$& $T_i \, [\mathrm{s}]$ & $k_i \, [\frac{\mathrm{rad}}{\mathrm{m}}]$ & $L_i \, [\mathrm{m}]$& $\epsilon_i \, [\mathrm{rad}]$ \\ 
\midrule
  1 &0.418  & 1.117 & 200 & $+0.50 \pi$  & 15.0 & 0.018 & 350.8 & 0.02 \\
2& 0.838 & 0.280 & 200 & $-0.20 \pi$ & 7.5 & 0.072 & 87.8 & 0.02 \\
3& 1.047 & 0.358 & 200 & $+0.75 \pi$ & 6.0 & 0.112 & 56.2 & 0.04\\
\bottomrule
\bottomrule
\end{tabular}
\label{tab:params_LWT}
\end{table}

According to the Nyquist-Shannon sampling theorem, at least two sampling points per wavelength are required $(\Delta x_\mathrm{wb}\le \tfrac{L}{2})$ \citep{Holthuijsen2007}. Hence, a  $\Delta x_\mathrm{wb}=25\, \mathrm{m}$ or $x_{\mathrm{wb},j}=\{0, 25, 50\} \, \mathrm{m}$, allows for detection of the shortest wave component with $L_3=56.2 \, \mathrm{m}$. To ensure deep water conditions $\tfrac{d}{L}>0.5$ also for the longest wave component with $L_1=350.8 \, \mathrm{m}$, the water depth is set to $d=200\, \mathrm{m}$. Moreover, selecting the wave periods as $T_i=\{6.0, \, 7.5, \, 15.0\}\, \mathrm{s}$ ensures that each wave component completes exactly five, four, or two periods within $30\, \mathrm{s}$. These aspects define the computational domain as 
\begin{equation}
\Omega= \big\{(x,t,z) \in \mathbb{R}^3 \big| 0 \le x \, [\mathrm{m}] \le 50, \;\; 0 \le t\, [\mathrm{s}] \le 30, \;\; -200 \le z \, [\mathrm{m}] \le \eta \big\}.
\end{equation}
The periodicity of the wave in the time interval allows using periodic boundary conditions in time
\begin{footnotesize}
\begin{align}
    \mathcal{L}_{\mathrm{PB},\eta} &= \frac{1}{N_\mathrm{p}} \sum_{i=1}^{N_\mathrm{p}} \left| \tilde{\eta}  \left( x_\mathrm{p}^{i} ,t_\mathrm{p}^{i}=t_0\right) - \tilde{\eta}  \left( x_\mathrm{p}^{i} ,t_\mathrm{p}^{i}=t_\mathrm{max}\right) \right|^2 \label{eq:PB_eta}\\
    \mathcal{L}_{\mathrm{PB},\Phi} &= \frac{1}{N_\mathrm{p}} \sum_{i=1}^{N_\mathrm{p}} \left| \tilde{\Phi}  \left( x_\mathrm{p}^{i} ,t_\mathrm{p}^{i}=t_0, z_\mathrm{p}^{i}\right) - \tilde{\Phi}  \left( x_\mathrm{p}^{i},t_\mathrm{p}^{i}=t_\mathrm{max}, z_\mathrm{p}^{i}\right) \right|^2\label{eq:PB_phi}
\end{align} 
\end{footnotesize}
for the analytic solution case in this section by adding $N_\mathrm{p}=1,000$ collocation points $\mathcal{P}_\mathrm{p}$. These terms are multiplied with adaptive weighting factors $\lambda_{\mathrm{PB}, \eta}$ and $\lambda_{\mathrm{PB}, \Phi}$ and added to the total loss in Eq. \eqref{eq:total_loss}.

\begin{figure}[h!]
\centering
\includegraphics[width=0.8\textwidth]{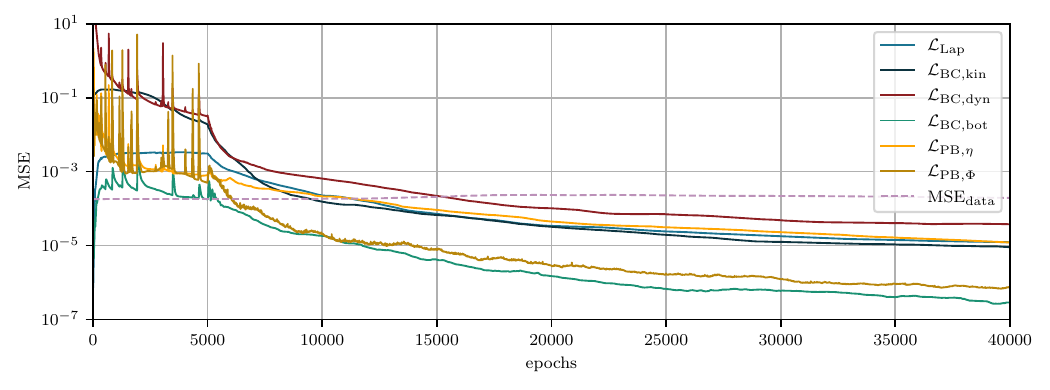}
\caption{Training loss curve of the PFT-PINN provided with sparse measurements from a superposition of three wave components for an assimilation task. The Adam optimizer for 5,000 epochs is followed by L-BFGS optimization. \textit{ReLoBRaLo} loss balancing ensures that each loss component $\mathcal{L}_i$ progresses with an approximately consistent rate relative to its initial MSE.}
\label{fig:loss_LWT}
\vspace{-0.3cm}
\end{figure}

We follow the training method detailed in Section \ref{sec:Method}, with the progress of the individual loss components $\mathcal{L}_\mathrm{i}$ over 5,000 epochs with the Adam optimizer and 35,000 epochs with the L-BFGS optimizer shown in Fig. \ref{fig:loss_LWT}. The desired behaviour of the \textit{ReLoBRaLo} loss balancing scheme (Sec.~\ref{sec:multi-objective optimization}) is evident: despite differences in measurement units and magnitudes, the optimizers balance the components, with each $\mathcal{L}_i$ progressing at a similar rate relative to its initial MSE. Moreover, the data error $\mathrm{MSE}_\mathrm{data}= \frac{1}{N_\mathrm{d}} \sum_{i=1}^{N_\mathrm{d}} \left| \eta_\mathrm{m}  \left( x_\mathrm{d}^{i} ,t_\mathrm{d}^{i}\right) - \tilde{\eta}  \left( x_\mathrm{d}^{i} ,t_\mathrm{d}^{i}\right) \right|^2 $ is plotted, confirming that the hard constraints are successfully learned during pre-training (Sec. \ref{sec:hard_constraints}).

\begin{figure}[h]
\centering
\includegraphics[width=0.88\textwidth]{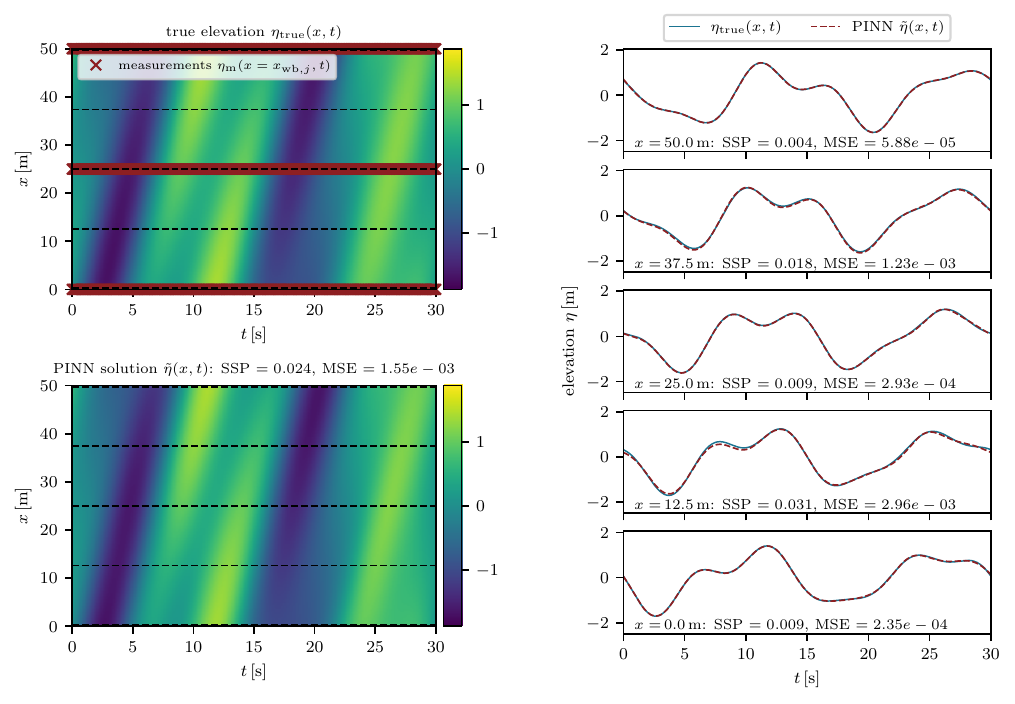}
\vspace{-0.2cm}
\caption{Ground truth surface elevation $\eta_\mathrm{true}(x,t)$ (upper left) from a superposition of three wave components, used to extract synthetic measurements $\eta_\mathrm{m}(x=x_{\mathrm{wb},j},t)$. The PFT-PINN's resulting solution for elevation $\tilde{\eta}(x,t)$ (lower left) closely matches the ground truth, as further confirmed by the alignment of $\eta_\mathrm{true}$ and $\tilde{\eta}$ in the cross-sections at five spatial points (right).}
\label{fig:LWT_ele}
\vspace{0.2cm}
\end{figure}

The success of reconstructing the sea state is shown in Figure \ref{fig:LWT_ele}, where the PINN's surface elevation solution $\tilde{\eta}(x,t)$ (lower left) is compared to the ground truth $\eta_\mathrm{true}(x,t)$ (upper left). Despite providing only three time-series measurements $\eta_\mathrm{m}(x=x_{\mathrm{wb},j}, t)$, the spatial elevation profile is accurately reconstructed in between, achieving an overall $\mathrm{SSP}=0.024$. The cross-section subplots (right) confirm this observation, although slightly higher errors occur between $\eta_\mathrm{m}$, e.g. a $ \mathrm{SSP}=0.031$ at $x=12.5 \, \mathrm{m}$.

\begin{figure}[h]
\centering
\includegraphics[width=0.88\textwidth]{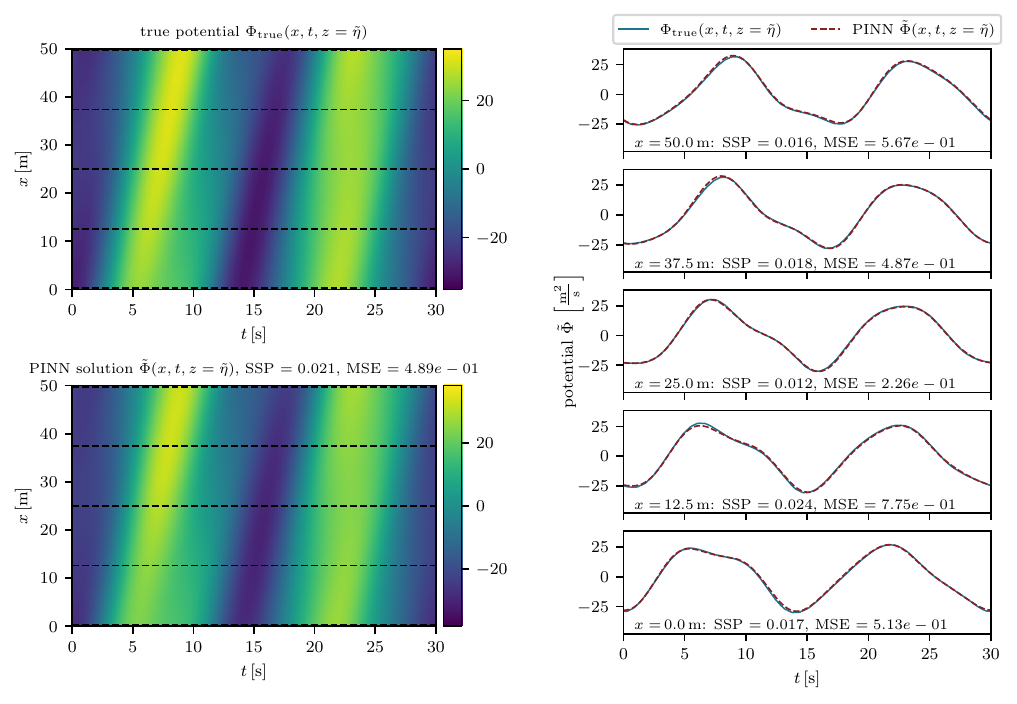}
\vspace{-0.2cm}
\caption{Ground truth potential $\Phi_\mathrm{true}(x,t, z=\eta_\mathrm{true})$ (upper left) from a superposition of three wave components. No potential measurements were provided for PFT-PINN training, its solution $\tilde{\Phi}(x,t, z=\tilde{\eta})$ (lower left) is entirely inferred from the coupling of $\tilde{\eta}$ and $\tilde{\Phi}$ in the surface BCs. Despite this, $\tilde{\Phi}$ closely matches $\Phi_\mathrm{true}$, as confirmed by the cross-section comparison (right).}
\label{fig:LWT_pot}
\end{figure}

Next, Figure \ref{fig:LWT_pot} compares the PINN's velocity potential evaluated at the surface $\tilde{\Phi}(x,t, z=\tilde{\eta})$ with the analytic ground truth $\Phi_\mathrm{true}(x,t, z=\eta_\mathrm{true})$. Notably, no direct potential measurements were provided for training, such that the PINN inferred it entirely from the coupling of $\tilde{\eta}$ and $\tilde{\Phi}$ in the free surface boundary condition loss components (Eqs. \eqref{eq:loss kin}-\eqref{eq:loss dyn}) and expanded it in depth direction via the Laplace equation (Eq. \eqref{eq:loss Laplace}) and the bottom BC component (Eq. \eqref{eq:loss bot}). Despite these challenges, the PINN's surface potential $\tilde{\Phi}$ closely aligns with the ground truth $\Phi_\mathrm{true}$, as also emphasised by a total error of $\mathrm{SSP}=0.021$. 

In summary, the results for using the PFT-PINN for wave data assimilation in this section support hypothesis~\ref{hyp:1}, by a successful reconstruction of an irregular wave elevation from sparse measurements. Moreover, they confirm hypothesis~\ref{hyp:2}, demonstrating that the corresponding velocity potential can be inferred solely from elevation data, without direct potential measurements.

\subsubsection{Nonlinear wave flume measurement in intermediate to deep water}
\label{sec:tank_assimilation}
 
As ocean waves are not only irregular but also inherently nonlinear, we now employ more complex real measurements for assimilation, stemming from laboratory wave flume experiments with high-resolution ground truth for wave elevation $\eta_\mathrm{true}(x,t)$ captured by cameras (Sec. \ref{sec:wave_tank_experiments}). In contrast to the LWT case in the previous section, no ground truth velocity potential is available for real data. Given the successful validation of potential reconstruction in the previous section, our method is still expected to remain reliable for nonlinear data, provided the PINN elevation $\tilde{\eta}(x,t)$ closely matches the reference $\eta_\mathrm{true}(x,t)$.

\begin{figure}[ht]
\centering
\includegraphics[width=0.8\textwidth]{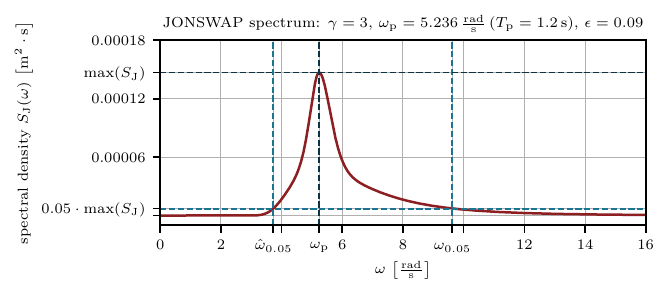}
\caption{JONSWAP spectrum used to generate the wave flume experiments. The horizontal line representing $5\%$ of the maximum spectral energy ($0.05 \cdot \mathrm{max}(S_\mathrm{J}))$ intersects the spectrum at $\omega_{0.05}=9.619 \, \frac{\mathrm{rad}}{\mathrm{s}}$. Consequently, we aim to capture and retain the wave components from the experiment for wavelengths down to its corresponding wavelength of $L_\mathrm{0.05}=0.666\, \mathrm{m}$. }
\label{fig:JONSWAP}
\end{figure}

Although spatio-temporal data $\eta_\mathrm{true}(x,t)$ is available, we mimick a less-equipped wave channel with only three buoys providing sparse time-series $\eta_\mathrm{m}(x=x_{\mathrm{wb},j}, t)$. To determine an appropriate buoy spacing for the experiment run specified by $T_\mathrm{p}=1.2 \, \mathrm{s}$, $\gamma=3$ and $\epsilon =0.09$, we aim to retain the wave components encompassing at least $5\%$ of the maximum energy in the JONSWAP spectrum, corresponding to $\omega_{0.05} = 9.619 \, \frac{\mathrm{rad}}{\mathrm{s}}$, as shown in Figure \ref{fig:JONSWAP}. This translates to a minimum wavelength of $L_{0.05} = 0.666 \, \mathrm{m}$ to be detectable. Following the Nyquist-Shannon sampling theorem $(\Delta x_\mathrm{wb} \leq \frac{L_{0.05}}{2})$ we determine the buoy spacing as $\Delta x_\mathrm{wb} = 0.3 \, \mathrm{m}$ and allow for the propagation of at least four peak periods, defining the computational domain
\begin{equation}
\Omega= \big\{(x,t,z) \in \mathbb{R}^3 \big| 0 \le x \, [\mathrm{m}] \le 0.6, \;\; 0 \le t\, [\mathrm{s}] \le 5, \;\; -0.7 \le z \, [\mathrm{m}] \le \eta \big\},
\end{equation}
where the $z$-component is constrained by the wave flume depth. A challenging irregular section is selected from the experiment, characterized by various wave heights and crest-trough asymmetries in $\eta_\mathrm{true}(x,t)$. While the wave component corresponding to $\omega_\mathrm{p}=5.24 \, \frac{\mathrm{rad}}{\mathrm{s}}$ has a length of $L_\mathrm{p}=2.17\, \mathrm{m}$ and with $\tfrac{d}{L_\mathrm{p}}=0.32$ is considered to travel in intermediate depth $\left(\tfrac{1}{20} \le \tfrac{d}{L} \le \tfrac{1}{2}\right)$, all components corresponding to $\omega \ge 6.62 \, \frac{\mathrm{rad}}{\mathrm{s}}$ in the JONSWAP spectrum have wavelengths $L \le 1.4 \, \mathrm{m}$, classifying them as deep-water waves $\left(\tfrac{d}{L} > \tfrac{1}{2}\right)$.

\begin{figure}[ht]
\centering
\includegraphics[width=0.8\textwidth]{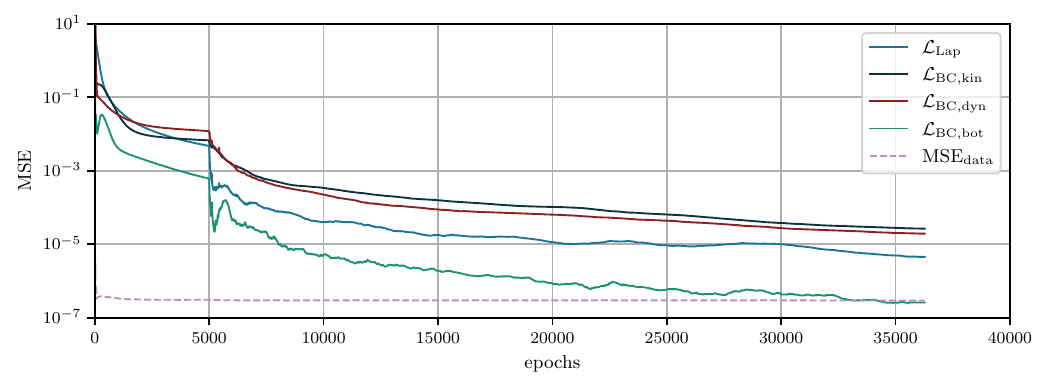}
\caption{Training loss curve of the PFT-PINN provided with sparse buoy measurements from the wave flume experiments for an assimilation task. The Adam optimizer is applied for 5,000 epochs, followed by L-BFGS optimization. Despite different wave amplitudes and domain size compared to Section \ref{sec:LWT_assimilation}, the curves again show successful convergence.}
\label{fig:loss_tank}
\vspace{0.1cm}
\end{figure}

The PINN again is trained as described in Sec. \ref{sec:Method}. Notably, periodic boundary conditions do not apply to wave tank data and the domain and wave amplitude are significantly smaller compared to the synthetic data in Sec. \ref{sec:LWT_assimilation}. Still, the loss curve in Fig. \ref{fig:loss_tank} converges well during training for 5,000 epochs using the Adam optimizer, followed by L-BFGS optimization terminating after approximately 36,000 epochs.

\begin{figure}[ht]
\centering
\includegraphics[width=0.88\textwidth]{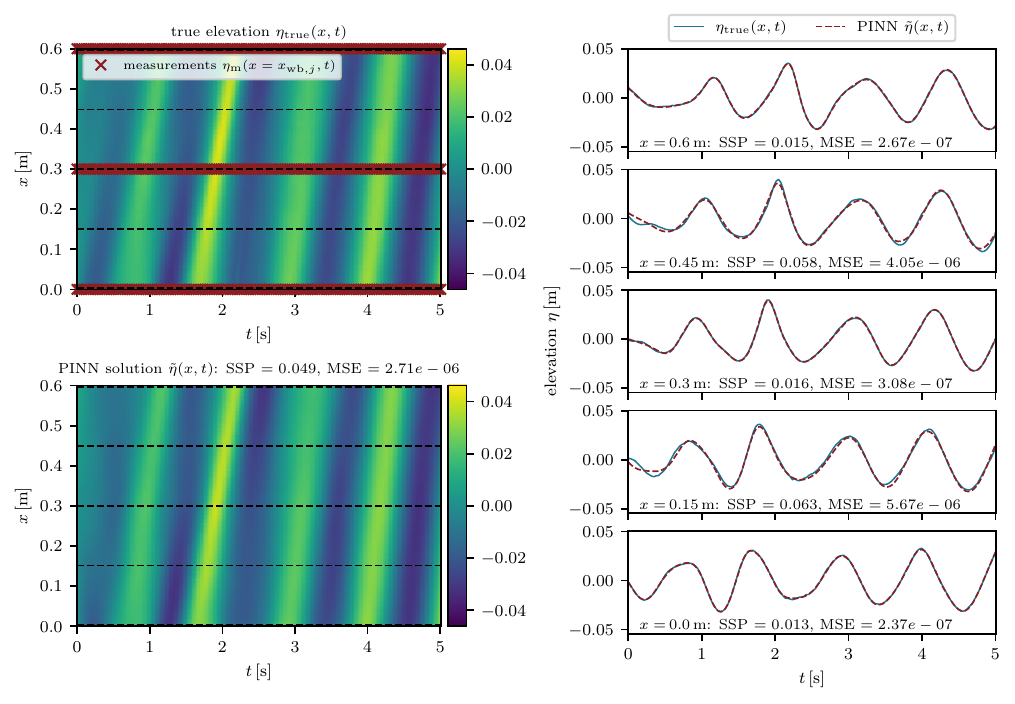}
\vspace{-0.2cm}
\caption{Ground truth surface elevation $\eta_\mathrm{true}(x,t)$ (upper left) from wave flume experiments. The PFT-PINN's reconstruction for elevation $\tilde{\eta}(x,t)$ (lower left) between the provided measurements $\eta_\mathrm{m}(x=x_{\mathrm{wb},j},t)$ closely matches the ground truth, as confirmed by the alignment of $\eta_\mathrm{true}$ and $\tilde{\eta}$ in the cross-sections at five spatial points (right).}
\label{fig:tank_ele}
\end{figure}

 The PINN reconstruction $\tilde{\eta}(x,t)$ in the lower-left of Figure~\ref{fig:tank_ele} shows strong agreement with the ground truth $\eta_\mathrm{true}(x,t)$ in the upper-left, as reflected by $\mathrm{SSP} = 0.049$. The cross-section subplots underline this observation, although the agreement at $x=0.15\, \mathrm{m}$ and $0.45\, \mathrm{m}$ is slightly weaker compared to the measurement points $x_{\mathrm{wb},j}$. Nevertheless, the SSP values remain in a highly satisfactory range below $0.1$. Furthermore, the training termination after 36,000 epochs suggests that the optimizer has reached its best outcome with the given hyperparameter configuration. Thus, the minor discrepancies between the ground truth $\eta_\mathrm{true}$ and the PINN solution $\tilde{\eta}$ may be attributed to real-world phenomena, that are not fully captured by the PFT, such as viscous effects from side-wall and bottom friction, for example.

\begin{figure}[ht]
\centering
\includegraphics[width=0.88\textwidth]{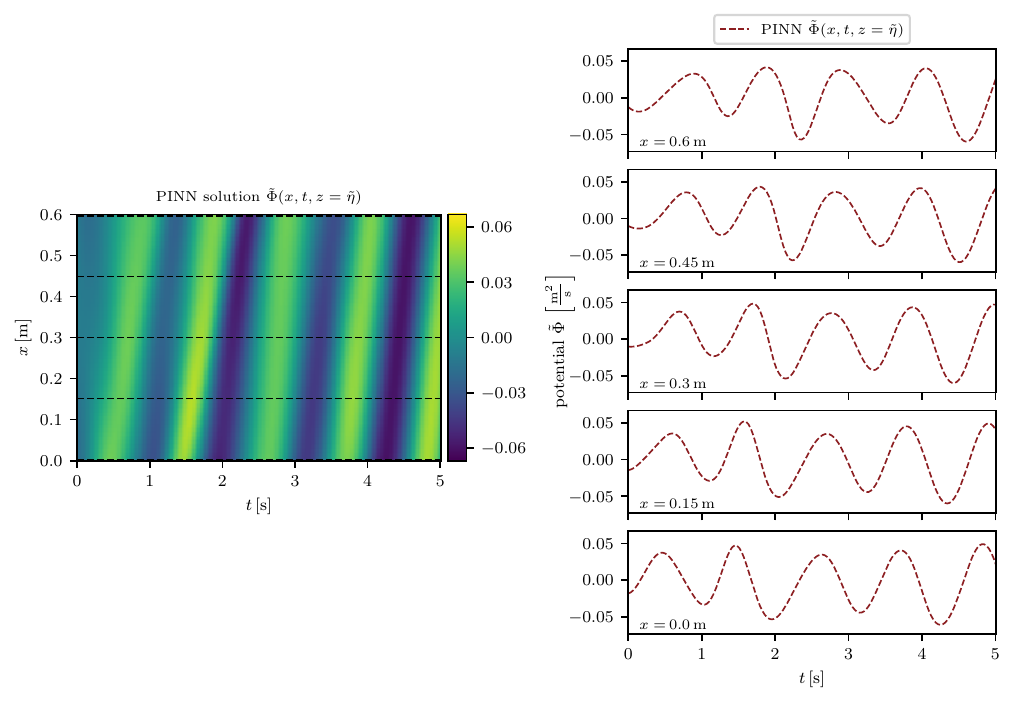}
\vspace{-0.2cm}
\caption{PFT-PINN's reconstruction of the potential on the free surface $\tilde{\Phi}(x,t,z=\tilde{\eta})$ for the wave flume experiments, that do not provide a ground truth for comparison. However, both potential amplitude and phase shift are within a reasonable range.}
\label{fig:tank_pot}
\end{figure}

Although there is no ground truth for the potential in the case of real data, the validity of the PINN's surface potential $\tilde{\Phi}(x,t,z=\tilde{\eta})$, illustrated in Figure \ref{fig:tank_pot}, can be roughly verified using the linear solution: The maximum wave elevation amplitude in Fig. \ref{fig:tank_ele} is $a_n \approx 0.04 \, \mathrm{m}$. Substituting this into Eq.~\eqref{eq:LWT_phi}, with $z=0 \, \mathrm{m}$ and $\omega_n = \tfrac{2 \pi}{T_\mathrm{p}}=5.24 \, \tfrac{\mathrm{rad}}{\mathrm{s}}$, yields a maximum potential amplitude of $\frac{g a_n}{\omega_n} \approx 0.07 \, \mathrm{m}$, which is consistent with $\tilde{\Phi}$ in Fig.~\ref{fig:tank_pot}. Additionally, we observe an approximate phase shift of $T_\mathrm{shift} \approx 0.3\, \mathrm{s}$ between the dominant wave crests (troughs) in Fig.~\ref{fig:tank_ele} and the dominant potential crests (troughs) in Fig.~\ref{fig:tank_pot}. Given the peak period of $T_\mathrm{p}=1.2\, \mathrm{s}$, this corresponds to a phase shift of $\frac{\pi}{2}$, further validating the approximate correctness of $\tilde{\Phi}$. Nonetheless, it is important to note, that this represents only a rough estimation, which we adopt based on the relatively low steepness of our measurement data. For real-world data, deviations from linear wave theory could be significantly larger.
In summary, this section revalidates hypotheses \ref{hyp:1} and \ref{hyp:2} and confirms hypothesis \ref{hyp:3}. The PINN assimilates non-linear, irregular wave surfaces between buoy measurements derived from a wave tank. The reconstruction achieves an error of $\mathrm{SSP} \le 0.1$, even for a particularly challenging surface elevation $\eta_\mathrm{true}(x,t)$.

\subsection{Prediction} 
\label{sec:prediction}

In the following, the wave prediction capabilities of the PFT-PINN are evaluated using wave flume measurements characterized by $T_\mathrm{p}=1.2\, \mathrm{s}$, $\gamma=3$, and $\epsilon=0.09$ from Sec.~\ref{sec:wave_tank_experiments}. Unlike the pure assimilation task relying on temporal buoy measurements in the last section, the prediction scenario assumes four spatial snapshots measurements $\eta_\mathrm{m}(x, t=t_{\mathrm{sn},k})$, similar to data obtained from radar systems. Thus, assimilation must occur between the sparse snapshots while simultaneously predicting the future wave evolution for $t>t_{\mathrm{sn},4}$. As we aim to propagate waves up to $\omega_{0.05}=9.619 \, \tfrac{\mathrm{rad}}{\mathrm{s}}$ (Fig. \ref{fig:JONSWAP}), corresponding to $T_{0.05}=0.653 \, \mathrm{s}$, the Nyquist-Shannon sampling theorem $\left( \Delta t_\mathrm{sn} \le \tfrac{T_{0.05}}{2}\right)$ and the camera's sampling ($\Delta t =0.0315\, \mathrm{s}$) yield a snapshot interval of $\Delta t_\mathrm{sn} =0.315 \, \mathrm{s}$. Moreover, each generated snapshot spans a range of $4\, \mathrm{m}$.

Besides, the concept of prediction region constrains the predictable spatio-temporal domain based on the group velocities $c_\mathrm{g} = \frac{\partial \omega}{\partial k}$ of the fastest and slowest components in a wave measurement \citep{Wu2004}. We select the limiting frequencies $\hat{\omega}_{0.05} =3.73 \, \tfrac{\mathrm{rad}}{\mathrm{s}}$ and $ \omega_{0.05}=9.62 \, \tfrac{\mathrm{rad}}{\mathrm{s}}$ from the spectrum in Fig. \ref{fig:JONSWAP} as intersections with the line representing $5\%$ of the maximal spectral energy. The corresponding wave numbers $\hat{k}_{0.05}$ and $k_{0.05}$ are determined by the dispersion relation. According to \cite{Naaijen2014}, linear wave theory provides a good approximation for the predictable region, with the limiting group velocities defined as
\begin{align}
    c_{\mathrm{g}, h} &=\left(\frac{1}{2} +\frac{\hat{k}_{0.05} d}{\sinh (2\hat{k}_{0.05} d)}\right)\cdot \frac{\hat{\omega}_{0.05}}{\hat{k}_{0.05}} =1.581 \, \frac{\mathrm{m}}{\mathrm{s}},\\
    c_\mathrm{g, \ell} &=\left(\frac{1}{2} +\frac{k_{0.05} d}{\sinh (2k_{0.05} d)}\right) \cdot \frac{\omega_{0.05}}{k_{0.05}} = 0.510 \, \frac{\mathrm{m}}{\mathrm{s}} 
\end{align}
for the selected measurement. For predicting one peak period into the future $(3 \cdot \Delta t_\mathrm{sn}+T_\mathrm{p})$, the set
 \begin{equation}
    \Omega =\big\{(x,t,z) \in \mathbb{R}^3 \big| c_{\mathrm{g},h} \cdot t -1.494 \le x \le c_\mathrm{g, \ell} \cdot t + 4, \;\; 0 \le t \le 2.145, \;\; -0.7 \le z \le \eta\big\},
\end{equation}
defines the computational domain for the prediction problem. 

The original PINN formulation \citep{Raissi2019} is a continuous-time approach, trained across all spatio-temporal collocation points simultaneously. While effective for assimilation, PINNs often struggle with forward problems, especially when solutions strongly depend on initial data \citep{Wang2024}. To handle the time-dependent nature of wave propagation, we adopt sequential training strategies \citep[cf.][]{Wight2020,Krishnapriyan2021,Mattey2022} by discretizing the time domain $[0, \, 3 \cdot \Delta t_\mathrm{sn}+T_\mathrm{p}] $ into $52$ segments
\begin{equation}
    \left[t_0= t_{\mathrm{sn},1}, \, t_1=t_{\mathrm{sn},4}\right], \, \left[t_1, \, t_2\right], \hdots, \left[t_{n-1}, \, t_n\right], \hdots, \left[t_{51}, \, t_{52}=t_{\mathrm{sn},4}+T_\mathrm{p}\right].
\end{equation}
The first time segment $[t_0, t_1]$ covers the entire assimilation period ($\Delta t_1 = 3 \cdot \Delta t_\mathrm{sn} = 0.945\, \mathrm{s})$ between the four wave snapshots $\eta_\mathrm{m}(x, t=t_{\mathrm{sn},k})$, while each subsequent segments $[t_{n-1}, \, t_n]$ spans $\Delta t_n =0.025 \, \mathrm{s}$. To ensure temporal causality, the PFT-PINN is initially trained using only the collocation points within the first segment to guarantee an accurate solution in this area. Every 750 training epochs, the collocation points of the next segment are added, allowing the solution to progressively evolve over time. This strategy mimics classical numerical methods, where a time discretization ensures the solution at each time step is resolved before approximating the solution at the subsequent time step \citep{Wang2024}.  

\begin{figure}[ht]
\centering
\includegraphics[width=0.8\textwidth]{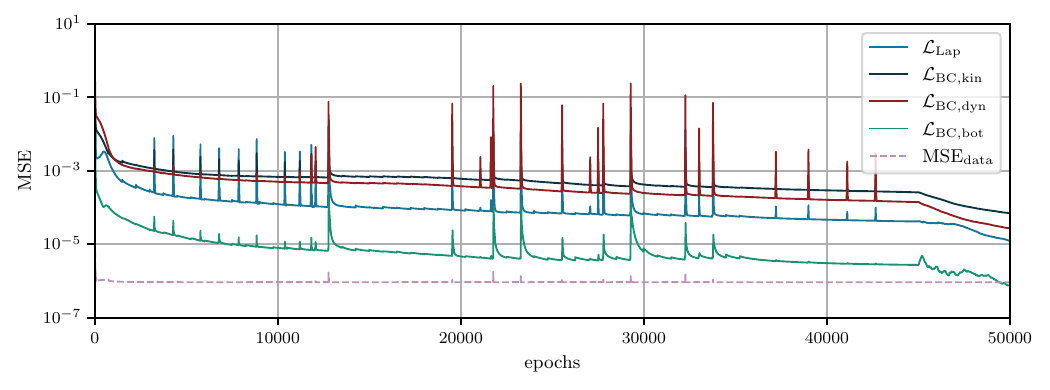}
\caption{Training loss curve for a prediction task of the PFT-PINN using sparse snapshot measurements from wave flume experiments. The Adam optimizer, applied for 45,000 epochs, is followed by L-BFGS optimization. The observed discontinuities every 750 epochs indicate the addition of collocation points for a new time segment ($\Delta t_n = 0.025 \, \mathrm{s}$) in the prediction region $t>t_{\mathrm{sn},4}$. These discontinuities may arise from fluctuations in gradient updates due to changes in the training data distribution as new collocation points are introduced. Despite some larger discontinuities, the successful convergence demonstrates the feasibility of predicting future wave dynamics.}
\label{fig:loss_prediction}
\end{figure}

Accordingly, for the prediction problem, training begins with $52 \cdot 750 =39,000$ epochs using the Adam optimizer, progressively expanding the time domain coverage. This is followed by refinement across the full spatio-temporal domain with $6,000$ additional Adam epochs and the L-BFGS optimizer, up to a total of $50,000$ epochs. An example loss curve is provided in Figure \ref{fig:loss_prediction}, where the discontinuities occurring every 750 epochs reflect the addition of collocation points from new time segments $[t_{n-1}, \, t_n]$ during training. 

\begin{figure}[ht!]
\centering
\includegraphics[width=0.88\textwidth]{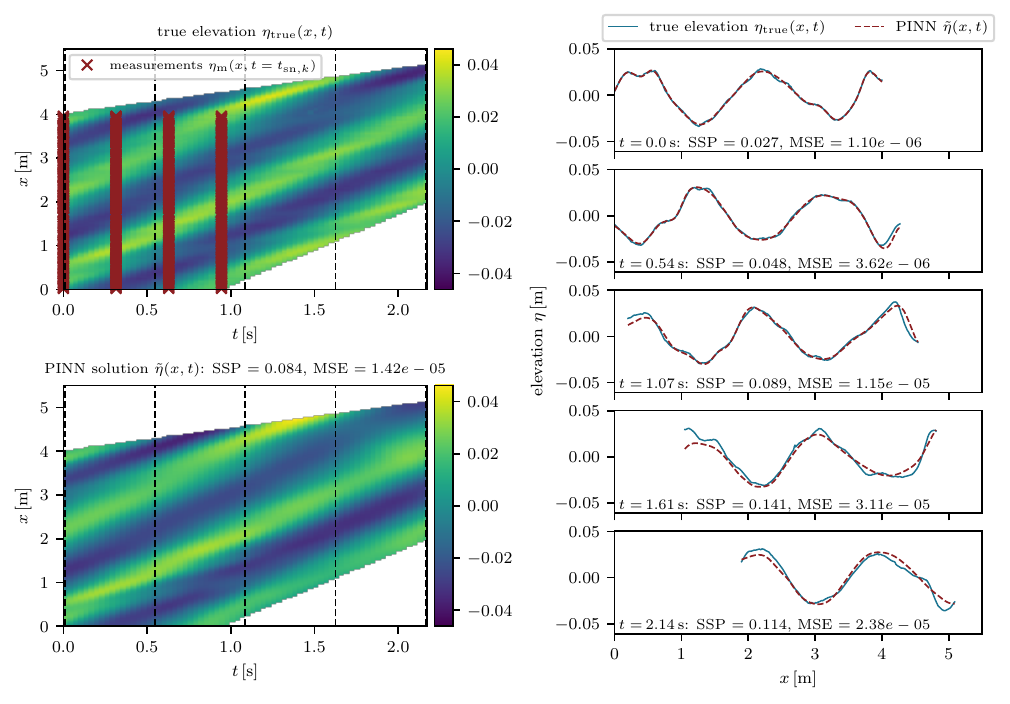}
\vspace{-0.2cm}
\caption{Ground truth surface elevation $\eta_\mathrm{true}(x,t)$ (upper left) from wave flume experiments in the prediction region limited by the min. and max. group velocity $c_{\mathrm{g},\ell}$ and $c_{\mathrm{g},h}$ on the top and bottom. Four snapshot measurements $\eta_\mathrm{m}(x,t=t_{\mathrm{sn},k})$ are provided for PINN training. The assimilation and prediction results $\tilde{\eta}(x,t)$ (lower left) closely matches the ground truth in the assimilation region $t\in [0, \, 0.945]\, \mathrm{s}$, as further confirmed by the first two cross-sections plots (right). As time progresses into the prediction region $t>0.945 \, \mathrm{s}$ the correspondence between $\eta_\mathrm{true}$ and $\tilde{\eta}$ is less strong, but still within an acceptable range.}
\label{fig:prediction_ele}
\end{figure}

The assimilation and prediction results corresponding to the loss curve, are presented in the lower-left subplot of Figure \ref{fig:prediction_ele}. The spatio-temporal elevation $\tilde{\eta}(x,t)$ is compared directly to the ground truth $\eta_\mathrm{true}(x,t)$ in the upper-left subplot, where the four snapshots $\eta_\mathrm{m}(x,t=t_{\mathrm{sn},k})$ provided for PFT-PINN training are marked. The boundaries of the prediction region, defined by the minimum and maximum group velocities, $c_{\mathrm{g},h}$ and $c_{\mathrm{g},\ell}$, are also visible, highlighting the PINN's capability to handle irregular computational domains. The ground truth and the PINN solution show an overall agreement characterized by $\mathrm{SSP}=0.084$.
More precisely, the snapshot cross-sections on the right side show strong agreement between the ground truth $\eta_\mathrm{true}$ and the PINN solution $\tilde{\eta}$ within the assimilation region $t \in [t_{\mathrm{sn},1}=0, \, t_{\mathrm{sn},4}=0.945 ] \, \mathrm{s}$.  However, as time progresses beyond $t_{\mathrm{sn},4}$, the accuracy decreases despite the use of sequential training strategies. This decline was expected given the highly dynamic nature of wave prediction and the well-known weaknesses of PINNs in forward problems \citep{Wang2024}. Nevertheless, even the $\mathrm{SSP}=0.141$ at $t=1.61\, \mathrm{s}$ can be considered acceptable, as the trends between the ground truth and prediction are still in close alignment (which this is also evident, for instance, at $\mathrm{SSP}=0.15$ in Fig.~\ref{fig:SSPs}). Moreover, despite the amplitudes are not entirely matched, the wave crests and troughs of $\eta_\mathrm{true}$ and $\tilde{\eta}$ are almost exactly phase-aligned, indicating that the PINN has accurately captured the wave dispersion for the prediction. Additionally, while not depicted here, the amplitudes and phase shifts of the PINN's potential $\tilde{\Phi}(x,t,z=\tilde{\eta})$ fall within expected ranges relative to the elevation data, in a similar manner as the observations in the last subsection in Fig.~\ref{fig:tank_pot}.

For this reason, we recognize many indications that support hypothesis \ref{hyp:4}, demonstrating that both assimilation and prediction can be addressed within the same framework, although the accuracy is generally less accurate than for the pure assimilation task.

\section{Conclusion and Outlook}
\label{sec:conclusion}

This study demonstrates that physics-informed neural networks (PINNs) are well-suited for integration with nonlinear potential flow theory (PFT) equations, commonly applied to reconstruct and predict ocean gravity waves. This compatibility arises from the \textit{universal approximation theorem}, stating that neural networks can approximate any nonlinear function. However, applying PFT in PINNs presents a particularly challenging case due to the requirement to solve for two coupled solution fields, velocity potential $\tilde{\Phi}(x,t,z)$ and surface elevation $\tilde{\eta}(x,t)$ within a dynamic, time-varying domain. Furthermore, measurements are sparse and typically available for only one solution component, usually the surface elevation.

To address these challenges, we train two neural networks in parallel, one for each solution field. Thereby, the output of the $\tilde{\eta}$-network informs the $z$-component of the $\tilde{\Phi}$-network to handle the time-varying free surface. 
Furthermore, Fourier feature embedding seems beneficial when handling multi-frequency wave data. Integrating sparse surface measurements as hard constraints in the $\tilde{\eta}$-network ensures accurate data adherence, enhancing training efficiency and preventing it from converging to neighbouring solutions. Additionally, the \textit{ReLoBraLo} loss balancing scheme addresses the multi-objective optimization challenge posed by various loss components of varying magnitudes.
We evaluate this PFT-PINN approach using three scenarios: 
\begin{itemize}[leftmargin=17pt]
\setlength{\itemsep}{0pt}
    \item[(1)] We demonstrated the assimilation of an irregular wave surface between sparse spatial measurements using an analytical solution of linear wave theory as ground truth surface elevation $(\eta_\mathrm{true})$ for extracting sparse synthetic measurements $(\eta_\mathrm{m})$ and comparing the PINNs reconstruction $(\tilde{\eta})$ post-training. This reconstruction is quantified by an error of $\mathrm{SSP}=0.024$, confirming hypothesis~\ref{hyp:1}. Analytical solutions additionally provide a ground truth potential $(\Phi_\mathrm{true})$, enabling the confirmation of hypothesis~\ref{hyp:2} by reconstructing the physically consistent velocity potential $(\tilde{\Phi})$ solely from surface elevation measurements $(\eta_\mathrm{m})$.
    \item[(2)] The second assimilation task involved sparse surface measurements $(\eta_\mathrm{m})$ from a laboratory wave flume, with additional high-resolution camera systems providing ground truth $(\eta_\mathrm{true})$ for comparing the PINN reconstruction $(\tilde{\eta})$. Since real waves contain nonlinear effects, the accurate reconstruction of $\mathrm{SSP}=0.049$ supports hypothesis~\ref{hyp:3}, indicating that the approach demonstrates excellent opportunities to investigate its performance using wave data with nonlinear and other effects incorporated. Although, real wave data do not provide a ground truth velocity potential $(\Phi_\mathrm{true})$ for an exact comparison, the amplitude, frequency and phase shift of the PINN's potential $(\tilde{\Phi})$ fall within expected ranges.
    
    \item[(3)] Finally, we demonstrated PFT-PINN's ability to predict nonlinear future wave dynamics while assimilating measurement data, confirming hypothesis~\ref{hyp:4}. This involved assimilation of a wave surface ($\tilde{\eta})$ from four sparse surface snapshots $(\eta_\mathrm{m})$ in $[t_{\mathrm{sn},1}, t_{\mathrm{sn},4}]$, while simultaneously predicting the wave dynamics one peak period into the future $(t>t_{\mathrm{sn},4})$. Compared to the ground truth ($\eta_\mathrm{true}$) the investigation yields an SSP of 0.084 averaged across the entire surface domain. However, the accuracy of the prediction decreases as the temporal distance from the last snapshot of the assimilation region increases.    
\end{itemize}

In summary, this study confirms that PFT-PINNs can capture nonlinear wave fields in assimilation and prediction tasks. Particularly, the method proves advantageous for pure assimilation tasks (\ref{step:2}), as indicated by the achieved SSP values in the first two scenarios. Moreover, existing numerical tools, already offer highly efficient solutions for wave prediction (\ref{step:3}) as long as initial conditions for the wave field are available. However, reconstructing these initial conditions from sparse measurements often requires extensive iterations using numerical tools, emphasising that the PFT-PINN approach could offer a promising alternative solution for wave data assimilation. 

Since this study demonstrates the feasibility of reconstructing $1\mathrm{D}+t$ wave surfaces $\eta(x,t)$, future work should extend the PFT-PINN approach to $2\mathrm{D}+t$ waves to include the y-direction. While this is expected to not involve great effort from an implementation point of view, additional wave propagation directions and their interactions in short-crested sea states could potentially affect reconstruction accuracy. Moreover, the higher dimension of $2\mathrm{D}+t$ data might increase training time significantly.
To facilitate real-time applications, improving the training speed of the PFT-PINN remains essential and could be achieved via code optimization, multi-GPU parallelization, or pre-training the PINN on general wave assimilation tasks, allowing rapid fine-tuning for new wave instances.
Moreover, while classical PINNs require retraining for each wave instance, exploring the physics-informed neural operator (PINO) approach \citep{Li2021b} offers a promising alternative to potentially generalize the reconstruction of sparse measurements to initial wave surfaces across various sea states.

With respect to future perspectives, we believe that potential flow PINNs might open up numerous novel applications in coastal or ocean engineering and oceanography. While numerical wave prediction methods are often constrained to regular numerical domains, the PFT-PINN can handle wave dynamics in highly irregular domains, such as bays, harbours, coastlines, etc.
Moreover, the PFT-PINN approach offers a straightforward way to integrate variable bottom topography $d(x)$ into the training.
The framework also allows extension for bathymetry inversion tasks, where bottom topography $d(x)$ is deduced from measured wave elevations $\eta_\mathrm{m}$, to generate water depth maps or assist the design of coastal infrastructure such as seawalls and breakwaters. 
Furthermore, the trained PFT-PINN provides inherently continuously differentiable solutions  $\tilde{\eta}(x,t)$ and $\tilde{\Phi}(x,t,z)$, enabling the evaluation of the potential at arbitrary points across the entire depth direction. This capability could support the analysis of fluid velocities near the seabed, valuable for studying sediment transport and erosion rates under varying wave conditions. Additionally, in offshore engineering, the knowledge of velocity potential across the entire depth facilitates the calculation of wave-induced pressure fields on submerged offshore structures.

\newpage

\section*{Funding}
This work was supported by the Deutsche Forschungsgemeinschaft (DFG - German Research Foundation) [project number 277972093: Excitability of Ocean Rogue Waves]. The laboratory work was funded by the UK Natural Environment Research Council (NERC) [grant number NE/T000309/1] awarded to AHC.
\section*{Acknowledgements}
The authors would like to express their gratitude to the lecturers George Em. Karniadakis and Khemraj Shukla, as well as to all organizers and funders for the Summer School on Physics-Informed Neural Networks and Applications at KTH, Stockholm in 2023.

\section*{Author Contributions}
\noindent Conceptualization: S.E, N.H. and M.S.\\
Data curation: A.H.C., R.C. and E.M.P. (wave experiments)\\
Formal analysis: S.E. (PINN), R.C. and Y.F. (wave experiments)\\
Funding acquisition: N.H. and A.H.C.\\
Investigation: S.E.\\
Methodology: S.E.\\
Project administration: N.H.\\
Resources: N.H. and A.H.C.\\
Software: S.E. (PINN), R.C. and Y.F. (image processing)\\
Supervision: N.H., T.T. and M.S.\\
Validation: S.E. and T.T.\\
Visualization: S.E.\\
Writing - original draft: S.E.\\
Writing - review and editing: N.H., T.T. A.H.C., R.C. and M.S.

\section*{Declaration of interests}
The authors declare that they have no known competing financial interests or personal relationships that could have appeared to influence the work reported in this paper.

\section*{Declaration of generative AI and AI-
assisted technologies in the writing process}

The investigations were conducted and the manuscript was written completely by the authors. Once the manuscript was completed, the authors used ChatGPT 3.5 (\url{https://chat.openai.com/}) in order to improve its grammar and readability. After using this tool, the authors reviewed and edited the content as needed and take full responsibility for the content of the publication.

\newpage

\appendix

\section{Graphical representation of SSP error metric}
\begin{figure}[ht!]
\centering
\includegraphics[width=0.88\textwidth]{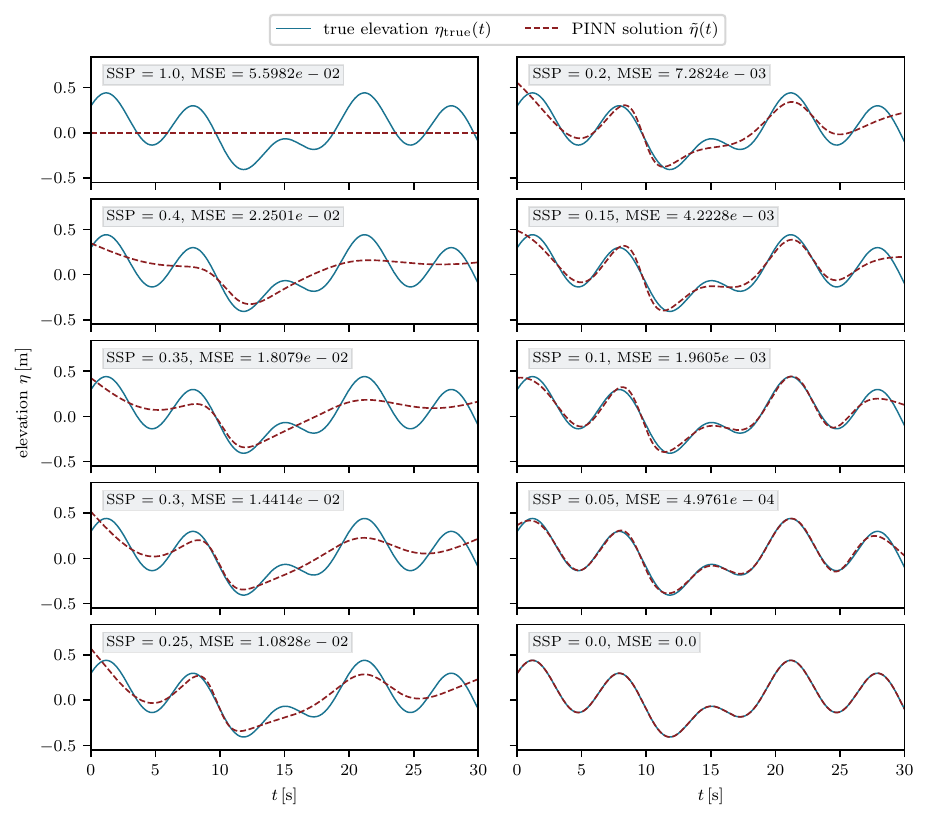}
      \caption{Graphical representation for assessing the meaning of different SSP values. The baseline signal $\eta_\mathrm{true}$ is compared to incrementally improving PINN solutions $\Tilde{\eta}$, resulting in decreasing SSP values. Additionally, MSE values are calculated to provide further insights.}
\label{fig:SSPs}
\end{figure}

\bibliography{apssamp}

\end{document}